\documentclass{article}
\usepackage{microtype}
\usepackage{graphicx}
\usepackage{subcaption}
\usepackage{booktabs} 

\usepackage{hyperref}
\usepackage{xspace}
\usepackage{multirow}
\usepackage{multicol}

\usepackage[preprint]{icml2026}

\usepackage{lineno}
\usepackage{amsmath}
\usepackage{amssymb}
\usepackage{mathtools}
\usepackage{amsthm}
\usepackage{pifont}

\usepackage[capitalize,noabbrev]{cleveref}

\theoremstyle{plain}

\theoremstyle{definition}

\theoremstyle{remark}

\usepackage[normalem]{ulem}

\usepackage[textsize=tiny]{todonotes}

\usepackage[table]{xcolor}
\usepackage{xcolor}
\definecolor{thomas}{RGB}{162, 82, 27}

\definecolor{dragonis}{RGB}{25, 86, 168}

\definecolor{add}{RGB}{192,23,36}

\makeatletter
\def\onedot{\ifx\@let@token.\else.\null\fi\xspace}
\def\eg{\emph{e.g}\onedot}
\def\ie{\emph{i.e}\onedot}
\makeatother

\begin{document}

\twocolumn[
  \icmltitle{Improving Reconstruction of Representation Autoencoder}

  \icmlsetsymbol{equal}{*}
  \icmlsetsymbol{pl}{$\dagger$}

  \begin{icmlauthorlist}
    \icmlauthor{Siyu Liu}{equal,nku,comp}
    \icmlauthor{Chujie Qin}{equal,nku}
    \icmlauthor{Hubery Yin}{pl,comp}
    \icmlauthor{Qixin Yan}{comp}
    \icmlauthor{Zheng-Peng Duan}{nku}
    \icmlauthor{Chen Li}{comp}
    \icmlauthor{Jing Lyu}{comp}
    \icmlauthor{Chun-Le Guo}{nku,nku_sz}
    \icmlauthor{Chongyi Li}{nku,nku_sz}
  \end{icmlauthorlist}

  \icmlaffiliation{nku}{VCIP, CS, Nankai University}
  \icmlaffiliation{comp}{WeChat Vision, Tencent Inc.}
  \icmlaffiliation{nku_sz}{NKIARI, Shenzhen Futian}
 
  \icmlcorrespondingauthor{Chongyi Li}{lichongyi@nankai.edu.cn}

  \icmlkeywords{Machine Learning, ICML}

  \vskip 0.3in
]




\printAffiliationsAndNotice{
\icmlEqualContribution $^\dagger$Project lead
}  

\begin{abstract}

Recent work leverages Vision Foundation Models as image encoders to boost the generative performance of latent diffusion models (LDMs), as their semantic feature distributions are easy to learn. 
However, such semantic features often lack low-level information (\eg, color and texture), leading to degraded reconstruction fidelity, which has emerged as a primary bottleneck in further scaling LDMs. 
To address this limitation, we propose LV-RAE, a representation autoencoder that augments semantic features with missing low-level information, enabling high-fidelity reconstruction while remaining highly aligned with the semantic distribution.
We further observe that the resulting high-dimensional, information-rich latent make decoders sensitive to latent perturbations, causing severe artifacts when decoding generated latent and consequently degrading generation quality. 
Our analysis suggests that this sensitivity primarily stems from excessive decoder responses along directions off the data manifold. 
Building on these insights, we propose fine-tuning the decoder to increase its robustness and smoothing the generated latent via controlled noise injection, thereby enhancing generation quality.
Experiments demonstrate that LV-RAE significantly improves reconstruction fidelity while preserving the semantic abstraction and achieving strong generative quality. Our code is available at \href{https://github.com/modyu-liu/LVRAE}{https://github.com/modyu-liu/LVRAE}.

\end{abstract}
    
\section{Introduction}
\label{sec:intro}

\begin{figure}[!t]
    \centering
    \includegraphics[width=\linewidth]{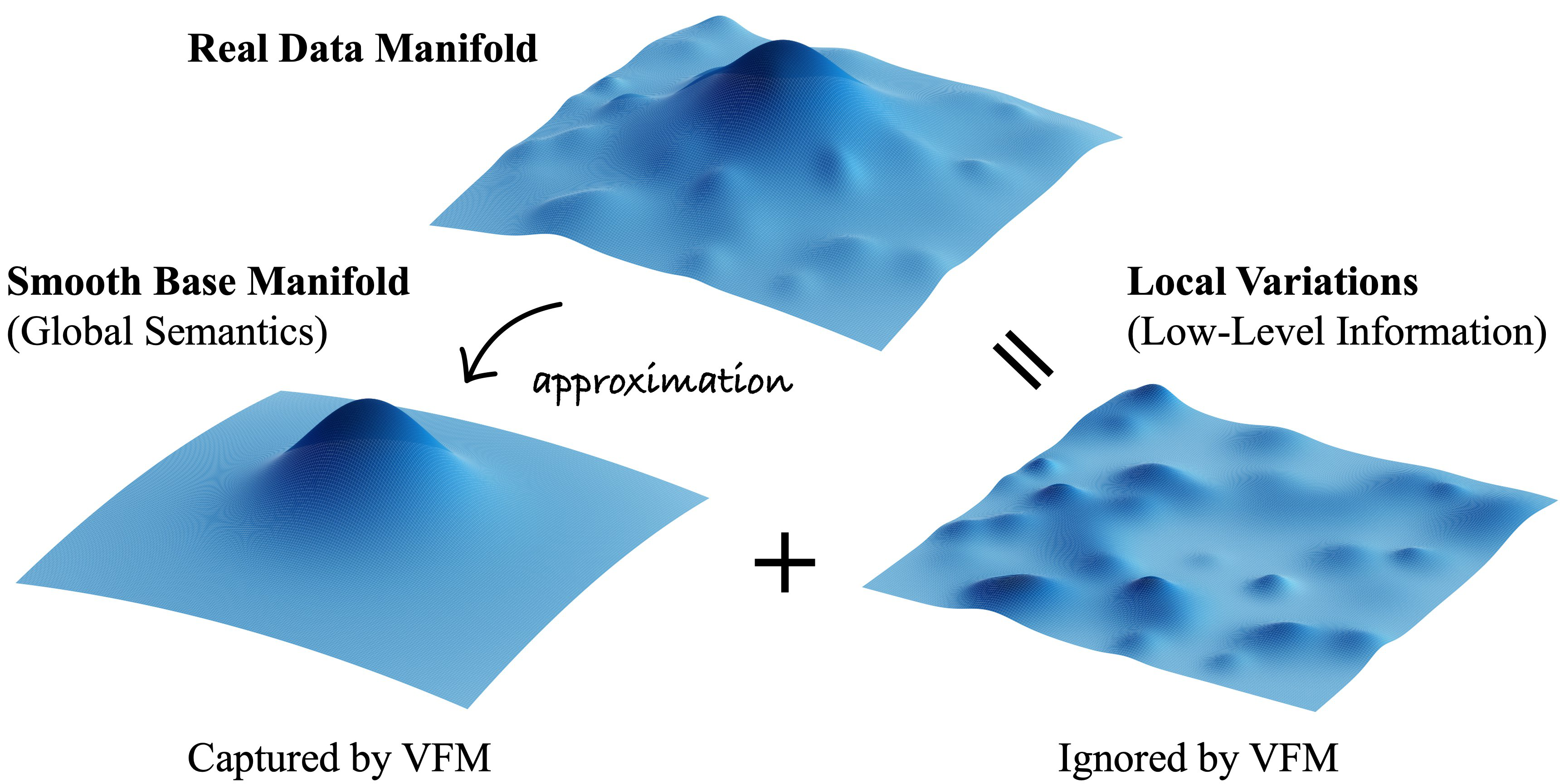}
    \caption{\textbf{Conceptual decomposition of the real data manifold.} We hypothesize that the real data manifold can be decomposed into two distinct components: a smooth base manifold representing global semantics (captured by VFMs) and local variations representing low-level information (ignored by VFMs). 
    }
    \label{fig:manifold}
    
    \vspace{-6mm}
\end{figure}

\begin{figure*}[!t]
    \centering
    \includegraphics[width=\textwidth]{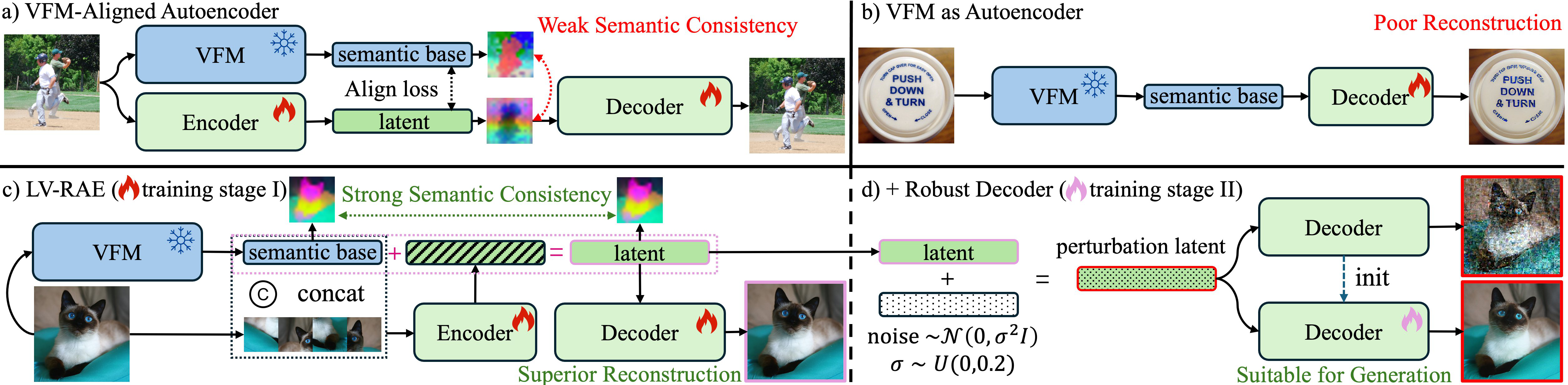}
    \caption{\textbf{Overview of previous methods and LV-RAE.}
    a) Training an autoencoder to align with VFMs fails to adequately preserve semantic consistency.
    b) Directly using a VFM as an autoencoder suffers from severely degraded reconstruction quality.
    c) The proposed LV-RAE significantly enhances reconstruction fidelity while effectively maintaining semantic representations.
    d) Fine-tune the LV-RAE decoder to improve robustness of latent perturbations, making it suitable for generation.
    }

    \label{fig:overview}
\end{figure*}

Variational Autoencoders (VAEs) \cite{kingma2013auto} serve as the foundation of latent diffusion models (LDMs) \cite{rombach2022high}, enabling diffusion processes to operate in a compact latent space while preserving visual fidelity. This paradigm has been instrumental in scaling diffusion models to high-resolution image synthesis.

Building on this paradigm, recent studies have further shown that leveraging Vision Foundation Models (VFMs) as feature encoders can substantially improve the generative performance of diffusion models, as their semantic features are linearly separable and easy for diffusion models to learn.
However, as shown in Fig.~\ref{fig:overview}(b), images reconstructed from VFM semantic features often suffer from degraded visual fidelity due to missing low-level information, limiting their applicability in tasks such as low-level vision \cite{liu2025faceme,chang2025pertouch} and precise image editing \cite{zhang2023adding,duan2025diffusion}.
More importantly, recent studies \cite{labs2025flux, esser2024scaling, team2025nextstep} indicate that reconstruction fidelity has emerged as a primary bottleneck when further scaling LDMs.

The fundamental issue stems from the inherent mismatch between what VFMs are trained to capture and what the features are required for faithful data reconstruction. 
Based on the manifold hypothesis \cite{carlsson2009topology}, as shown in Fig.~\ref{fig:manifold}, we consider the real data manifold as a low-dimensional manifold that can be decomposed into a smooth base manifold for global semantics and local variations for low-level information. 
Since semantic features focus on high-level understanding, they capture the smooth base but ignore the local variations. This explains why VFM-based reconstructions often lack visual fidelity. 

To bridge this gap, a straightforward solution is to fine-tune the VFM with a reconstruction loss to fit local variations, alongside an alignment loss to preserve the base manifold structure. 
However, this strategy can be sub-optimal. 
The two losses introduce competing objectives, causing the base manifold itself to continuously drift during training and depriving the local variations of a stable reference. 
As a result (\cref{exp:ab_ft}), the learning of local variations becomes unstable and fails to converge to a consistent solution. 

To address this limitation, we propose the \textbf{L}ocal-\textbf{V}ariations Augmented \textbf{R}epresentation \textbf{A}uto\textbf{e}ncoder (LV-RAE). As shown in Fig.~\ref{fig:overview}(c), rather than forcing the latent to directly align with semantic features, LV-RAE treats the semantic features as a fixed base manifold, and employs a shallow encoder to learn the missing low-level information (i.e., local variations) not captured by the VFM. Specifically, the encoder takes both the input image and its corresponding semantic features as inputs. The decoder then reconstructs the image by adding the encoder outputs to the semantic features.
Notably, we find that only minimal adjustments to the semantic features are required to achieve high-fidelity image reconstruction (PSNR$\sim$32.32) while remaining highly aligned with semantics (CKNNA$\sim$0.99).

Furthermore, as shown in Fig.~\ref{fig:overview}(d), we observe that decoders become increasingly sensitive to latent perturbations as the latent dimensionality increases. 
We attribute this sensitivity to excessive decoder responses along off-manifold directions. 
In higher-dimensional latent spaces, more such directions exist, allowing even small deviations to accumulate and be amplified by the decoder.
To mitigate this issue, we propose a dual-stage approach. 
First, we fine-tune the decoder with stochastic latent noise to regularize its local response and enhance its inherent robustness. 
Second, during the diffusion sampling phase, we inject controlled noise into the generated latent to dynamically modulate the decoder’s effective gain, thereby suppressing off-manifold artifacts and improving overall generation quality.

Our contributions are summarized as follows:
\begin{itemize}
    \item We propose LV-RAE, an improved representation autoencoder that achieves high-fidelity reconstruction while preserving strong semantic alignment.
    \item We analyze decoder sensitivity in high-dimensional latent spaces and introduce a strategy to improve decoder robustness for improved generation.
    \item Extensive experiments demonstrate that LV-RAE produces diffusion-friendly latent and achieves impressive generative quality. 
    
\end{itemize}

\section{Related Work}

\subsection{Autoencoders for Latent Diffusion Models}

Autoencoders are a core component of LDMs, as their design directly determines the quality and efficiency of the models. 
Prior studies have primarily focused on improving the compression ratio~\cite{chen2024deep, chen2025dc} and reconstruction fidelity~\cite{labs2025flux, esser2024scaling}. 
Recent studies suggest that the structure of the latent space, \ie, \textit{diffusability}  \cite{skorokhodov2025improving}, plays a critical role in diffusion model training. 
To improve diffusability, some studies \cite{kouzelis2025eq,skorokhodov2025improving} have observed that latent representations often contain excessively high-frequency components and have proposed simple regularization strategies to suppress these components.
In parallel, a line of research \cite{yao2025reconstruction,leng2025repa,xiong2025gigatok} has explored aligning latent with semantic features extracted from VFMs \cite{oquab2023dinov2,simeoni2025dinov3}, demonstrating substantially faster convergence and improved generative performance. 
Despite their success, these methods often suffer from an information bottleneck when aligning compact latents with high-dimensional VFM features. 
This constraint makes it challenging for the latent space to simultaneously accommodate high-level semantic priors and fine-grained low-level information, thereby both degrading reconstruction fidelity and hindering further performance gains.

\subsection{Representation Autoencoders}
To further enrich the semantic density of the latent space, recent studies \cite{zheng2025diffusion,shi2025latent,chen2025aligning,bi2025vision,gao2025one} have explored the direct use of VFMs as encoders.
Traditionally, diffusion models are thought to struggle with learning high-dimensional latent distributions.
Consequently, several works~\cite{chen2025aligning, bi2025vision, gao2025one} have attempted to distill high-dimensional semantic features into lower-dimensional latents while preserving semantic information. 
In contrast, RAE~\cite{zheng2025diffusion} demonstrates that diffusion models can be trained directly on high-dimensional semantic features and achieve strong generative performance with only minimal architectural modifications.
Nevertheless, a key limitation remains: VFM features are primarily optimized for high-level tasks and lack the low-level information required for image reconstruction. 
As a result, image reconstruction based on such semantic features often suffers from noticeable visual degradation.

\section{Method}

In this section, we first introduce the proposed LV-RAE (\cref{sec:3.1}).
We then analyze the decoder’s sensitivity to latent perturbations through a toy experiment (\cref{sec:3.2}), and propose an augmentation strategy to enhance decoder robustness (\cref{sec:3.3}).
The overall framework is illustrated in Fig.~\ref{fig:overview} (c) and (d).

\subsection{Improving Representation Autoencoder}\label{sec:3.1}

\paragraph{Motivation}
As illustrated in Fig.~\ref{fig:overview}, our goal is to address the poor reconstruction fidelity when directly using VFMs as encoders. VFM features typically reside in extremely high-dimensional spaces, \ie, with channel dimensions far exceeding those of conventional VAE latents. We argue that in such high-dimensional semantic spaces, even a minimal change can lead to substantial improvements in reconstruction fidelity.

A straightforward approach is to finetune the VFM using both a reconstruction loss and an alignment loss. However, we find that this strategy is sub-optimal. Such a training paradigm implicitly forces the encoder to simultaneously fit the base manifold and the local variations defined on top of it. Since the base manifold itself keeps evolving during training, the corresponding local variations lack a stable reference and therefore cannot converge consistently toward a well-defined direction, ultimately hindering effective optimization.

To overcome this limitation, we propose the Local-Variations Augmented Representation Autoencoder (LV-RAE), which departs from this alignment paradigm. Instead of forcing the encoder to extract both semantic and low-level information into a shared latent, we treat the VFM semantic features as a fixed base manifold. The encoder is then tasked solely with learning the low-level information that is missing from the semantic features, leading to more stable and effective optimization. 

\paragraph{Architecture}
To facilitate training and maintain consistency with the VFM framework, both the encoder and decoder of our autoencoder adopt a Transformer architecture equipped with RoPE \cite{su2024roformer} positional embeddings, using a patch size of $16 \times 16$. The encoder consists of 6 Transformer layers and is designed to be lightweight, focusing on learning low-level information. The decoder is deeper, comprising 12 Transformer layers, to effectively reconstruct high-fidelity images.

\begin{figure}
    \centering
    \includegraphics[width=\linewidth]{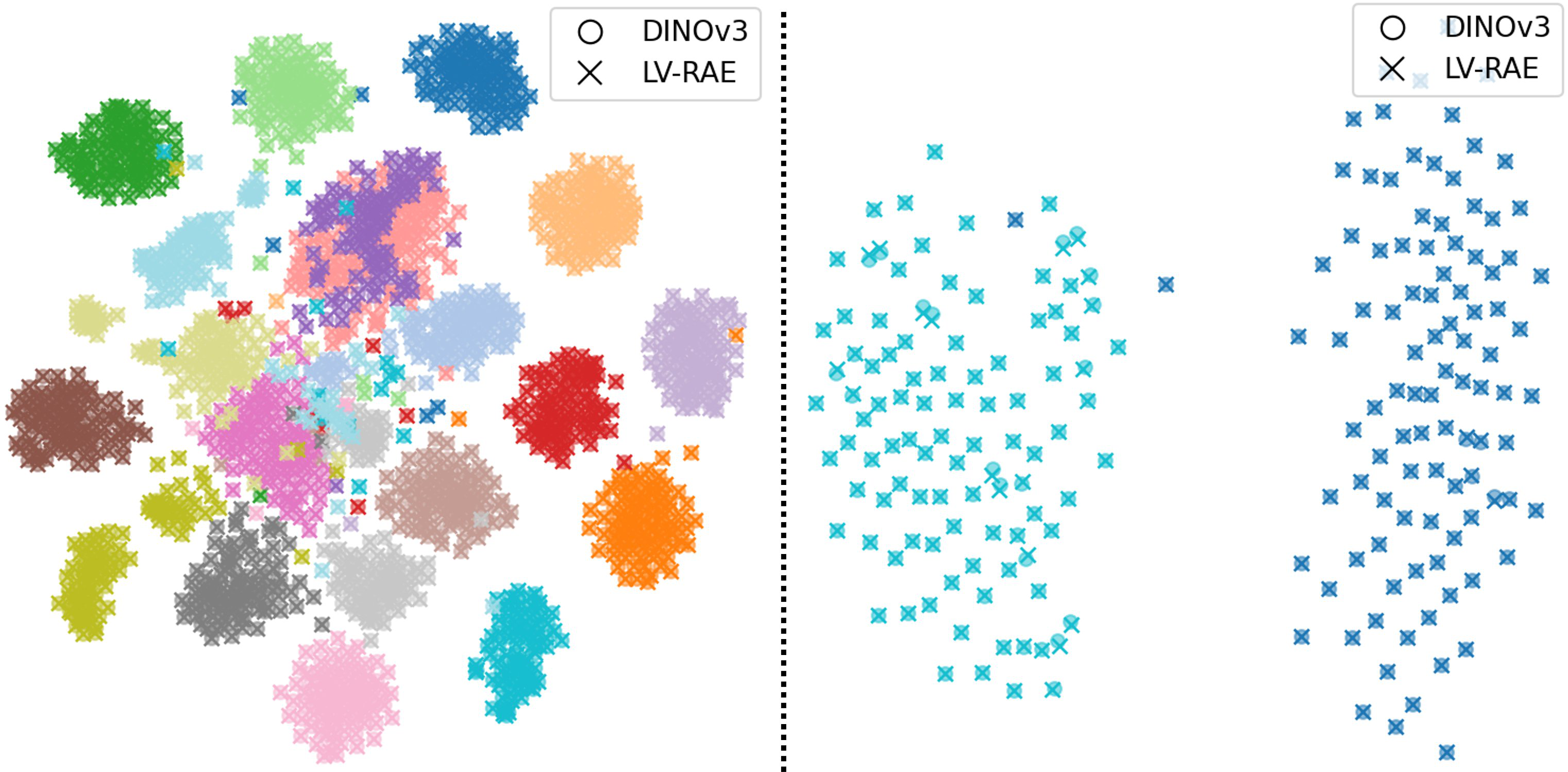}
    \caption{\textbf{The t-SNE visualization of DINOv3's semantic features with LV-RAE's latents in a shared representation space.} \textbf{Left}: 20-class setting. \textbf{Right}: 2-class setting. LV-RAE latents exhibit strong overlap with DINOv3 semantic features, suggesting that they lie in a tightly shared representation space with minimal distributional discrepancy. }
    \label{fig:t-sne}
\end{figure}

\paragraph{Training} 
Let $\Phi$ denote a VFM, \ie, DINOv3 \cite{simeoni2025dinov3}, and let $E(\cdot)$ and $D(\cdot)$ denote the encoder and decoder, respectively.
Given an image $X\in \mathbb{R}^{H\times W \times 3}$, the VFM $\Phi$ extracts semantic features $u \in \mathbb{R}^{N\times D}$. Notably, $u$ is obtained before the final LayerNorm of $\Phi$.

The image $X$ is first projected into a patch-level representation $x_{\text{in}} \in \mathbb{R}^{N \times D}$ via a $16 \times 16$ convolution, where $N = \frac{H}{16}\times\frac{W}{16}$. The projected features are then concatenated with the semantic features $u$ along the token dimension, forming
\begin{equation}
x = [x_{\text{in}}; u] \in \mathbb{R}^{2N \times D},    
\end{equation}
which serves as the input to the encoder $E$. Let $r = E(x)[:N,:] \in \mathbb{R}^{N \times D}$ denote the first $N$ tokens of the encoder output. We interpret these tokens as encoding the low-level information missing from the semantic features. The complete latent  $z\in \mathbb{R}^{N\times D}$ is then obtained by element-wise addition of $r$ and $u$, followed by a LayerNorm:
\begin{equation}
    z=\text{LayerNorm}(r+u).
\end{equation}
Notably, we initialize the final linear layer of the encoder with all-zero weights, and initialize the LayerNorm using the parameters from the final LayerNorm of the VFM to avoid disrupting semantic features at initialization.

The decoder $D$ then reconstructs the image from $z$, producing $\bar{X}\in \mathbb{R}^{H\times W \times 3}$. 
For image reconstruction, we employ a combination of the pixel-wise loss $\mathcal{L}_{1}$ and the perceptual loss $\mathcal{L}_{lpips}$~\cite{johnson2016perceptual,zhang2018unreasonable}:
\begin{equation}
\mathcal{L}_{rec}=\alpha\mathcal{L}_{1}(X,\bar{X})+\beta\mathcal{L}_{Lpips}(X,\bar{X}),
\end{equation}
where $\alpha=\beta=1$ are weighting coefficients.
Simultaneously, to ensure the learned latent space remains grounded in the semantic manifold, we explicitly align the latent representation $z$ with the semantic features $u$ via an $\mathcal{L}_2$ loss:
\begin{equation}
\mathcal{L}_{\text{align}} = \| z - u \|_{2}^{2}.
\end{equation}

The final training objective is:
\begin{equation}
\mathcal{L} = \mathcal{L}_{\text{rec}} + \eta \cdot \mathcal{L}_{\text{align}},
\end{equation}
where $\eta=5$ is the weighting coefficient. As shown in Fig.~\ref{fig:t-sne}, the latents produced by LV-RAE closely align with the corresponding VFM semantic features, indicating that they reside in a shared representation space.

\begin{figure}[t]
    \centering
    \includegraphics[width=\linewidth]{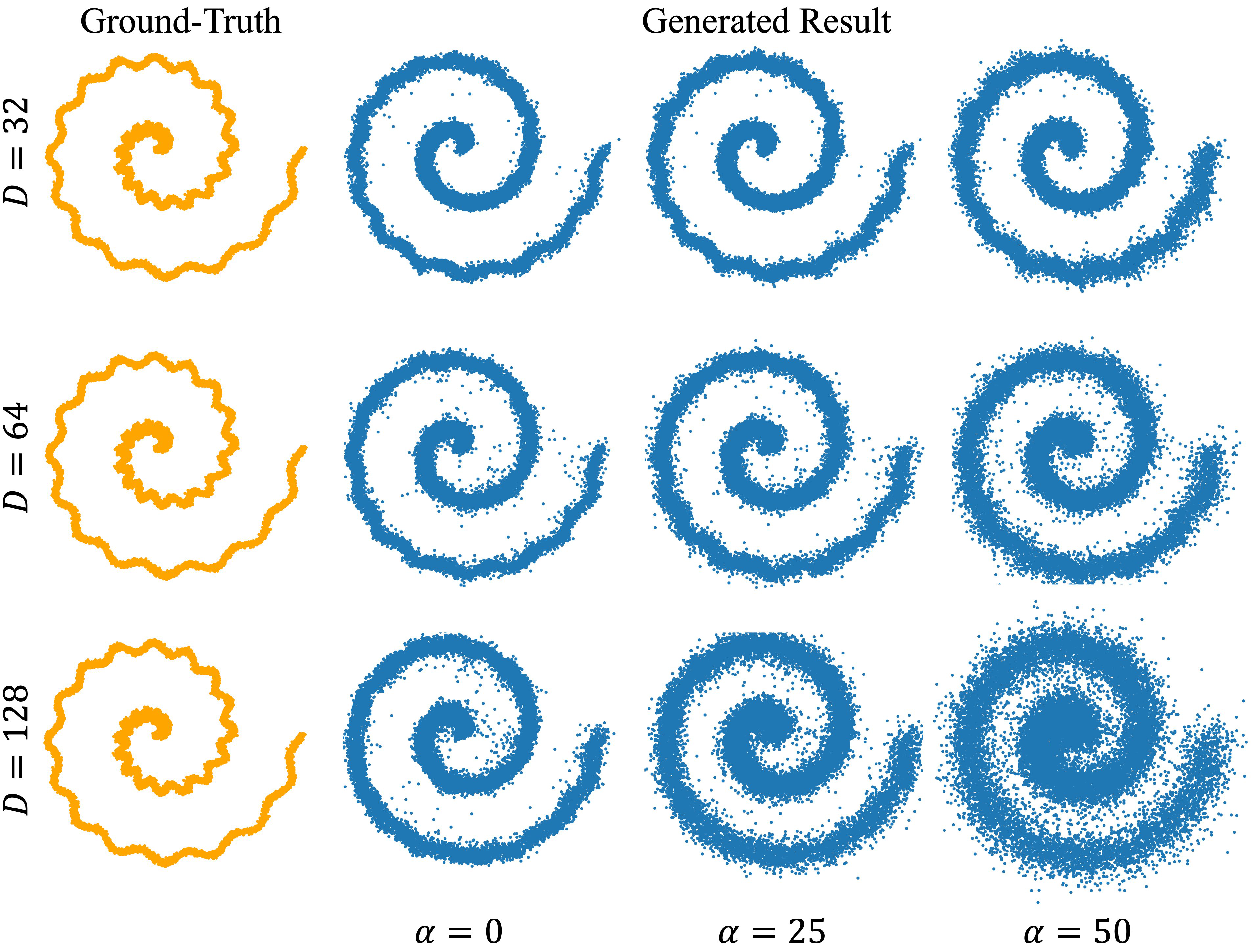}
    \caption{\textbf{Toy Experiment.} A 2-dimensional underlying data is embedded into a $D$-dimensional space to train a diffusion model. The generated samples are projected back to two dimensions using a decoder that responds to both manifold and off-manifold directions for visualization. The parameter $\alpha$ controls the decoder’s sensitivity to off-manifold directions. In the high-dimensional setting ($D=128$), increasing the decoder’s sensitivity to off-manifold directions (larger $\alpha$) causes deviations along these directions to accumulate and be amplified by the decoder, resulting in severe departures from the ground truth.}
    \label{fig:toy_exp}
\end{figure}

\begin{figure}[t]
    \centering
    \includegraphics[width=\linewidth]{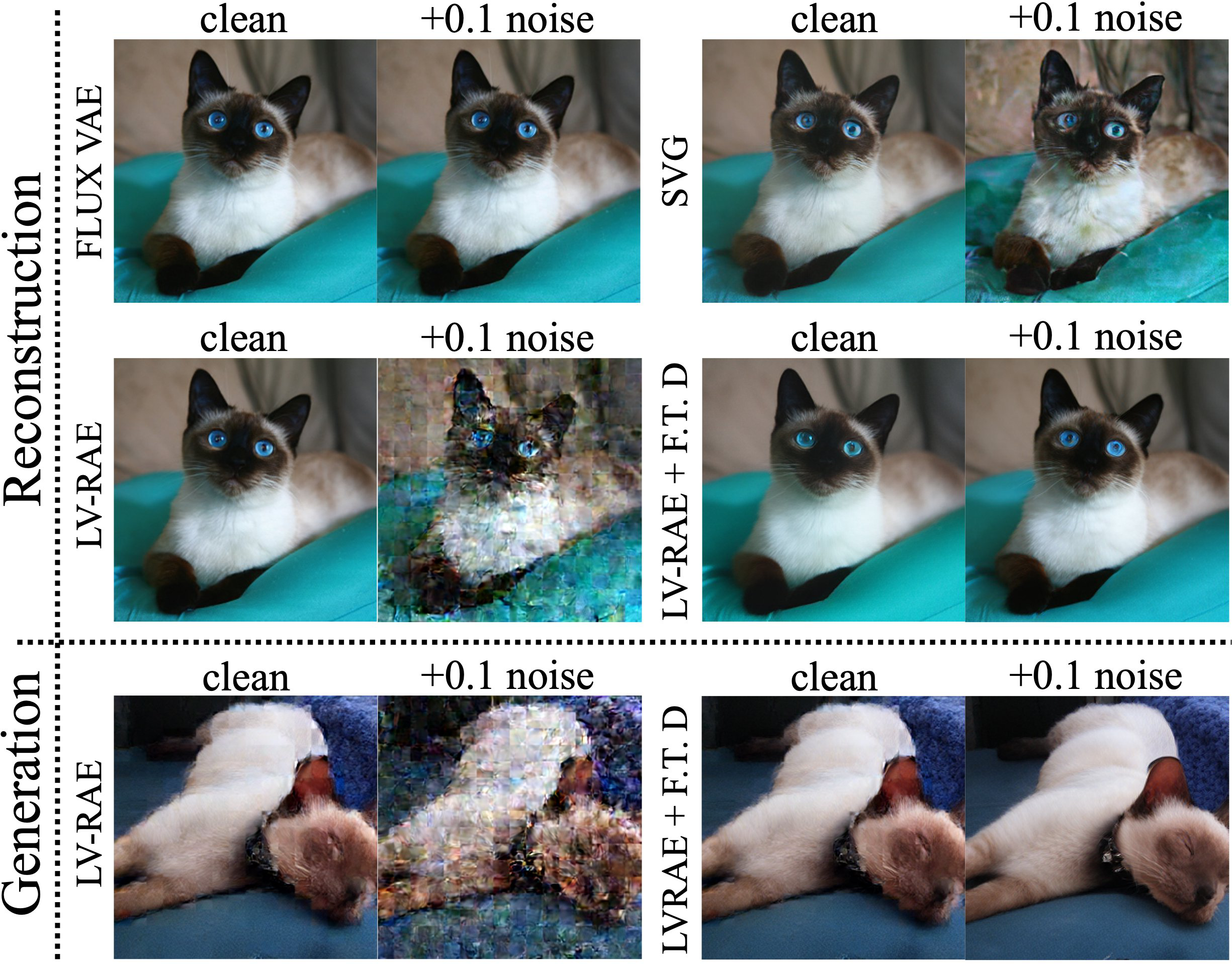}
    \caption{\textbf{Visualization of decoder sensitivity to latent perturbations.} \textbf{Top:} In high-dimensional latent spaces, the decoder is sensitive to latent perturbations, even small perturbations (e.g., +0.1 noise) can produce pronounced pixel-level artifacts. \textbf{Bottom:} This sensitivity undermines generation because generative models often struggle to accurately capture the true data distribution, making even minor sampling shifts capable of causing significant structural distortions in the output. Our approach enhances generation quality by fine-tuning the decoder to increase robustness to latent perturbations and smoothing the generated latent via controlled noise injection.}
    \label{fig:noise_exp}
\end{figure}

\subsection{Toy Experiment}\label{sec:3.2}

Despite the high reconstruction fidelity of LV-RAE, as shown in Fig.~\ref{fig:noise_exp}, we observe that its decoder is exceptionally sensitive to perturbations in the latent space.
Under the manifold assumption, real data distributions typically lie on low-dimensional manifolds. Mapping such low-dimensional manifolds into a high-dimensional space inevitably introduces many directions orthogonal to the data manifold. Along these directions, the decoder can exhibit excessively large Jacobian magnitudes, causing even small perturbations in the latent space lead to noticeable visual artifacts in the reconstructed outputs. We verify this assumption via a toy experiment.

We begin by sampling points $\hat{x} \in \mathbb{R}^2$ from a simple distribution and embedding them into a $D$-dimensional latent space via a random orthonormal projection matrix $P \in \mathbb{R}^{D \times 2}$, where $P^{\top}P = I_{2}$. The resulting latents are denoted by $z = P \hat{x} \in \mathbb{R}^D$. A simple diffusion model, \ie, 5-layer ReLU MLP with 512-dimensional hidden units, is then trained to model this distribution in the latent space.

To explicitly investigate the effect of the decoder, we construct a decoder $D(\cdot)$ that is designed to respond to latent directions both parallel and orthogonal to the data manifold. Specifically, let $U \in \mathbb{R}^{D \times (D-2)}$ be an orthonormal basis whose columns are orthogonal to those of $P$, \ie, $P^{\top}U=0$. The decoder is formulated as:
\begin{equation}
    D(z) = P^{\top}z  + \alpha \sin(\beta U^{\top}z ) W,
\end{equation}
where $\alpha$ and $\beta$ are hyperparameters, $W  \in \mathbb{R}^{(D-2) \times 2}$ is a random projection matrix with entries independently sampled from $\mathcal{N}(0,\frac{1}{D-2}\mathrm{I})$. To formally characterize the decoder's sensitivity, we derive its Jacobian matrix:
\begin{equation}
    J_D(z) = \frac{\partial D(z)}{\partial z}= P^{\top} + \alpha \beta W^{\top}\mathrm{Diag}(\cos(\beta U^{\top}z))U^{\top}.
\end{equation}
The second term controls the decoder’s response to perturbations along directions orthogonal to the data manifold through the total gain factor $\alpha \beta$. 
For simplicity, we set $\beta = 1$, such that the overall gain of the off-manifold component is determined solely by $\alpha$.

We conduct experiments under different latent dimensionalities and varying values of $\alpha$, with the results shown in Fig.~\ref{fig:toy_exp}. When $\alpha = 0$, decoding reduces to a pure projection onto the data-manifold directions. In this regime, increasing the latent dimensionality does not induce significant changes in the generated distribution. The overall manifold structure remains stable with only minor variations.

In contrast, when $\alpha > 0$ and the decoder is allowed to respond to off-manifold directions, the effect of latent dimensionality becomes pronounced. As the dimensionality increases, the generated samples exhibit progressively larger deviations from the ground-truth distribution, eventually leading to severe distortions and loss of structure at large $\alpha$. This behavior arises because higher-dimensional latent spaces provide more off-manifold directions, along which even small deviations can accumulate and be amplified by the decoder.

These results verify that controlling the decoder’s response along off-manifold directions is essential for stable generation, especially in high-dimensional latent spaces. 

\subsection{Decoder with Noise Augmentation}\label{sec:3.3}

\paragraph{Motivation}

To suppress the decoder's excessive responses, a simple yet effective solution is to fine-tune the decoder with latent noise. From a Jacobian perspective, noise injection implicitly penalizes excessive local gain, reducing the decoder’s sensitivity along all directions. 

Since the approximation accuracy of diffusion priors is model-dependent, a fixed decoder can't adapt to all diffusion models. We propose a decoupled framework that transforms decoder sensitivity into a tunable parameter. Specifically, we introduce random noise injection during training to encourage robustness across varying levels of latent uncertainty. At inference time, additional noise is injected into the generated latent to dynamically adjust the decoder’s effective response, allowing it to accommodate different degrees of manifold misalignment introduced by the diffusion process.

\paragraph{Training}
Building upon the LV-RAE trained in the previous stage, instead of feeding the original latent $z$ into the decoder, we input a noise-perturbed latent $\tilde{z}$ obtained by adding Gaussian noise:
\begin{equation}
\tilde{z} = z + \sigma \cdot \epsilon, \quad \sigma \sim \mathcal{U}(0, \tau),\quad \epsilon \sim \mathcal{N}(0,\mathrm{I}), 
\end{equation}
where $\tau=0.2$ is a hyperparameter for controlling the maximum noise magnitude.
The decoder then reconstructs the image from the perturbed latent: $ \bar{X} = D(\tilde{z})$.

In addition to using the reconstruction loss $\mathcal{L}_{rec}$, we incorporate an adversarial loss \cite{goodfellow2014generative} with an adaptive weighting strategy. Specifically, we compute the gradients
of the reconstruction loss $\mathcal{L}_{rec}$ and the GAN loss $\mathcal{L}_{gan}$
with respect to the decoder’s final layer:
\begin{equation}
w_{\text{gan}} = 
\frac{\left\| \nabla \mathcal{L}_{\text{rec}} \right\|}
     {\left\| \nabla \mathcal{L}_{\text{gan}} \right\|}.
\end{equation}
The overall training objective is:
\begin{equation}
    \mathcal{L}=\mathcal{L}_{rec}+\kappa \cdot w_{\text{gan}} \cdot \mathcal{L}_{\text{gan}},
\end{equation}
where $\kappa=0.75$ is the weighting coefficient. 

\paragraph{Sampling}
At the generation stage, we apply noise injection to the final latent representation produced by the generative model. Concretely, given a generated latent $z_0$, we perturb it with Gaussian noise of a fixed magnitude:
\begin{equation}
\tilde{z}_0 = z_0 + \bar{\sigma} \cdot \epsilon, \quad \epsilon \sim \mathcal{N}(0, \mathrm{I}),
\end{equation}
where $\bar{\sigma}$ controls the noise strength at inference time, larger values result in smoother outputs, as empirically validated in \cref{sec:ab_noise_for_generation}. The final image is then obtained by decoding the perturbed latent: $X = D(\tilde{z}_0)$.

\begin{table*}[!t]
    \centering
    \caption{\textbf{Quantitative comparison of reconstruction performance across different autoencoders.} Our proposed LV-RAE achieves state-of-the-art reconstruction quality among all autoencoders. Notably, under nearly lossless semantic preservation, our method significantly outperforms SVG in terms of reconstruction quality, demonstrating its superiority.}
    \resizebox{\textwidth}{!}{
    \begin{tabular}{lccccccccc}
    \toprule
        \multirow{2}{*}{Tokenizer}
        & \multirow{2}{*}{Config.}
        & \multicolumn{4}{c}{ImageNet-1K}
        & \multicolumn{4}{c}{COCO 2017} \\
        \cmidrule(lr){3-6} \cmidrule(lr){7-10}
        &  & PSNR$\uparrow$ & SSIM$\uparrow$ & LPIPS$\downarrow$ & rFDD$\downarrow$
        & PSNR$\uparrow$ & SSIM$\uparrow$ & LPIPS$\downarrow$ & rFDD$\downarrow$ \\
    \midrule
        SD-VAE \cite{rombach2022high}        & f8d4  &  25.65 &  0.748 & 0.068 & 13.78 & 25.38  & 0.760 & 0.065 & 43.72\\
        FLUX-VAE \cite{labs2025flux} & f8d16 & 31.01 & 0.914 & \textbf{0.017} & 2.88 & 30.89 & 0.921 & \textbf{0.015} & 8.86 \\ 
        VA-VAE \cite{yao2025reconstruction}        & f16d32  & 26.26 & 0.786 & 0.046 & 4.18 & 26.07 & 0.799 & 0.044 & 18.76 \\
        SVG \cite{shi2025latent}        & f16d392  & 21.87 & 0.632 & 0.114 & 24.92 & 21.71 & 0.645 & 0.112 & 60.66 \\
        \rowcolor{gray!15}
        LV-RAE (\textcolor{blue!70}{ours})    & f16d768 & \textbf{32.50} & \textbf{0.938} & \textbf{0.017} & 5.51 & \textbf{32.32} & \textbf{0.941} & \textbf{0.015} & 13.20 \\
        \rowcolor{gray!15}
        + Noise Augmentation (\textcolor{blue!70}{ours})   & f16d768 & 31.11 & 0.927 & 0.019 & \textbf{2.49} & 30.91 & 0.930 & 0.018 & \textbf{8.74} \\
       
    \bottomrule
    \end{tabular}
    }
    \label{tab:reconstruction_result}
\end{table*}

\begin{table}[!t]
    \centering
    \vspace{-3mm}
    \caption{\textbf{Reconstruction quality and semantic alignment comparison between fine-tuning VFMs and the proposed LV-RAE.} Across all VFMs, LV-RAE consistently outperforms fine-tuning the VFM in terms of both reconstruction quality and semantic alignment. In particular, when using DINOv3 as the backbone, LV-RAE achieves remarkably strong reconstruction performance while maintaining a high degree of semantic alignment.}
    \resizebox{\linewidth}{!}{
    \begin{tabular}{lccccc}
    \toprule
    Method & Target VFM & Init. D. & PSNR$\uparrow$ & SSIM$\uparrow$ & CKNNA$\uparrow$ \\
    \midrule
    
    \multirow{2}{*}{F.T. VFM} & \multirow{2}{*}{DINOv3-B} & \ding{55} & 29.94 & 0.899 & 0.925 \\
     &  & $\checkmark$ & 30.88 & 0.920 & 0.925 \\
    \rowcolor{gray!15}
    LV-RAE & DINOv3-B & \ding{55} & \textbf{32.32} & \textbf{0.941} & \textbf{0.987} \\ 
    
    \midrule
    F.T. VFM & DINOv2-B & \ding{55} & 25.58 & 0.770 & 0.873 \\
    \rowcolor{gray!15}
    LV-RAE & DINOv2-B & \ding{55} & \textbf{29.30} & \textbf{0.892} & \textbf{0.982} \\ 

    \midrule
    F.T. VFM & SigLip2 & \ding{55} & 23.48 & 0.683 & 0.802  \\
    \rowcolor{gray!15}
    LV-RAE & SigLip2 & \ding{55} & \textbf{27.66} & \textbf{0.840} & \textbf{0.989} \\

    \bottomrule
    \end{tabular}
    }
    \label{tab:ab_lvrae}
\end{table}

\begin{figure*}[t]
    \centering
    \includegraphics[width=\textwidth]{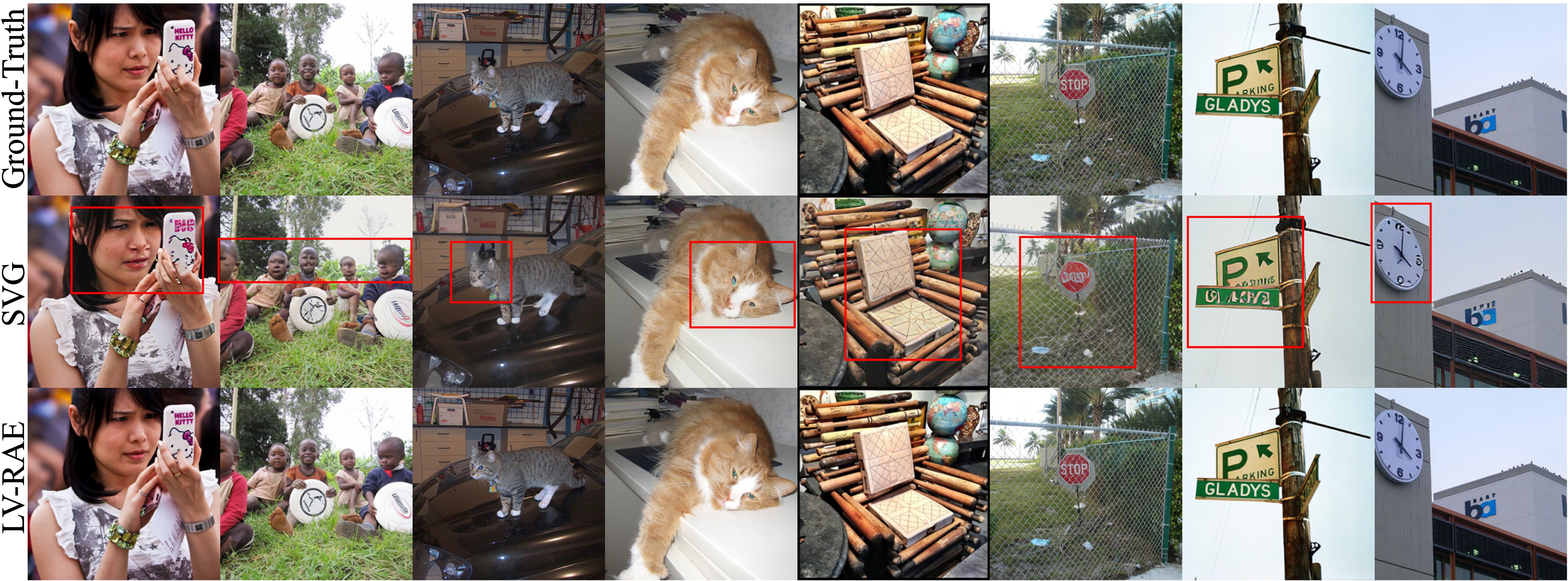}
    \caption{\textbf{Qualitative comparison of reconstruction fidelity with SVG.} While SVG preserves color consistency with the ground truth, it suffers from severe structural distortions, particularly in regions containing text, human faces, animal faces, symbols, and grid patterns. These distortions significantly degrade perceptual quality. In contrast, the proposed LV-RAE effectively handles these challenging structures and produces visually faithful reconstructions. Critical regions are highlighted using red boxes. \textbf{Zoom in for best view.}}
    \label{fig:rec_result_vis}
    
\end{figure*}

\begin{figure}[!t]
    \centering
    \vspace{-3mm}
    \includegraphics[width=\linewidth]{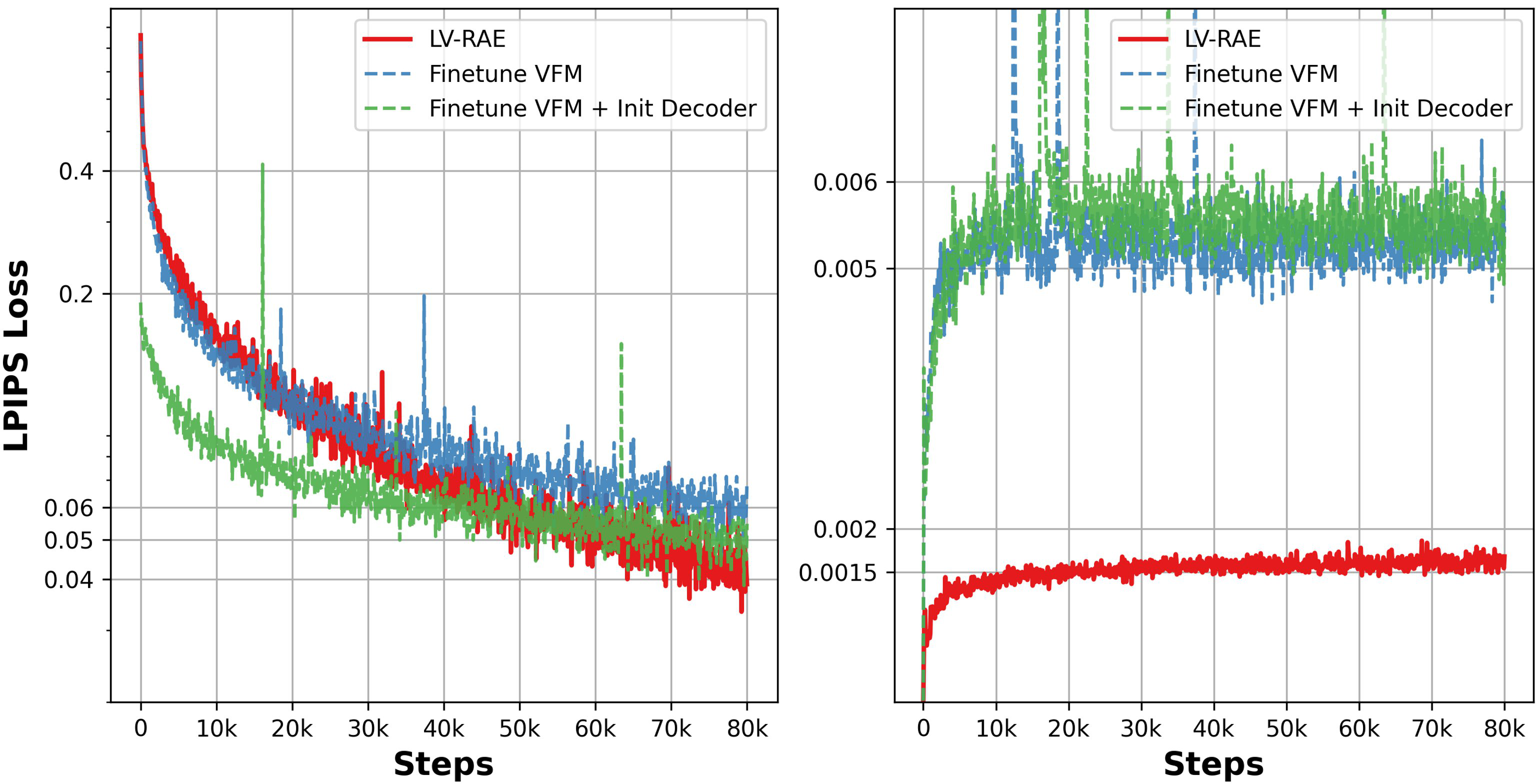}
    \caption{\textbf{Training dynamics comparison of different methods.} \textbf{Left:} LPIPS loss over training steps. \textbf{Right:} Semantic alignment loss over training steps. The proposed LV-RAE achieves a lower and more stable alignment loss throughout training and converges to a lower LPIPS loss compared with directly fine-tuning the VFM.
    }
    \label{fig:train_loss}
\end{figure}

\begin{table*}[!t]
    \centering
    \caption{\textbf{Class-conditional generation performance on ImageNet 256×256.}}
    \resizebox{\textwidth}{!}{
    \begin{tabular}{lcccccccccc}
    \toprule
        \multirow{2}{*}{\textbf{Method}}
        &  \multirow{2}{*}{\textbf{\#Params}} & \multirow{2}{*}{\textbf{Epochs}}
        & \multicolumn{4}{c}{\textbf{Generation@256 w/o guidance}}
        & \multicolumn{4}{c}{\textbf{Generation@256 w/ guidance}} \\
        \cmidrule(lr){4-7} \cmidrule(lr){8-11}
        &  & & \textbf{gFID}$\downarrow$ & \textbf{IS}$\uparrow$ & \textbf{Prec.}$\uparrow$ & \textbf{Rec.}$\uparrow$
         & \textbf{gFID}$\downarrow$ & \textbf{IS}$\uparrow$ & \textbf{Prec.}$\uparrow$ & \textbf{Rec.}$\uparrow$ \\
        
    \midrule
        \multicolumn{11}{l}{\textit{\textbf{Pixel Diffusion}}} \\ 
    \midrule
        ADM \cite{dhariwal2021diffusion}  & 554M & 400 & 10.94 & 101.0 & 0.69 & 0.63 & 3.94 & 215.8 & 0.83 & 0.53 \\
        RIN \cite{jabri2022scalable} & 410M & 480 & 3.42 & 182.0 & - & - & - & - & - & - \\
        PixelFlow \cite{chen2025pixelflow} & 677M & 320 & - & - & - & - & 1.98 & 282.1 & 0.81 & 0.60 \\
    \midrule
        \multicolumn{11}{l}{\textit{\textbf{Latent Diffusion}}} \\ 
    \midrule
        MaskDiT \cite{Zheng2024MaskDiT} & 675M & 1600 & 5.69 & 177.9 & 0.74 & 0.60 & 2.28 & 276.6 & 0.80 & 0.61 \\
        DiT \cite{Peebles2022DiT} & 675M & 1400 & 9.62 & 121.5 & 0.67 & 0.67 & 2.27 & 278.2 & \textbf{0.83} & 0.57 \\
        SiT \cite{ma2024sit} & 675M & 1400 & 8.61 & 131.7 & 0.68 & 0.67 & 2.06 & 270.3 & 0.82 & 0.59 \\
        FasterDiT \cite{yao2024fasterdit} & 675M & 400 & 7.91 & 131.3 & 0.67 & \textbf{0.69} & 2.03 & 264.0 & 0.81 & 0.60 \\
        REPA \cite{yu2025repa} & 675M & 800 & 5.90 & 157.8 & 0.70 & \textbf{0.69} & 1.42 & 305.7 & 0.80 & 0.65 \\
        VA-VAE \cite{yao2025reconstruction} & 675M & 800 & \textbf{2.17} & 205.6 & \textbf{0.77} & 0.65 & 1.35 & 295.3 & 0.79 & 0.65 \\
        DDT \cite{wang2025ddt} & 675M & 400 & 6.27 & 154.7 & 0.68 & \textbf{0.69} & \textbf{1.26} & \textbf{310.6} & 0.79 & 0.65 \\
    SVG \cite{shi2025latent} & 675M & 1400 & 3.36 & 181.2 & - & - & 1.92 & 264.9 & - & - \\   
    \midrule  
        \multicolumn{11}{l}{\textit{\textbf{Latent Diffusion with LV-RAE + 0.1 noise (Ours)}}} \\ 
    \midrule
        \rowcolor{gray!15}
        DiT-XL & 675M & 400 & 3.77 & 185.37 & 0.76 & 0.60 & - & - & - & - \\
    \midrule
        \rowcolor{gray!15}
        $\text{DiT}^{\text{DH}}$-XL &  839M & 800 & 2.42 & \textbf{223.8} & \textbf{0.77} & 0.64 & 1.82 & 249.7 & 0.75 & \textbf{0.67} \\
    \bottomrule
    \end{tabular}
    }
    \label{tab:generation_result}
\end{table*}

\section{Experiments}

\subsection{Experiments Setup}

\paragraph{Training} 
We train all models on the ImageNet-1K \cite{russakovsky2015imagenet} $256\times256$ dataset. For generation, we use LightningDiT \cite{yao2025reconstruction} and $\text{DiT}^{\text{DH}}$ \cite{zheng2025diffusion} as backbones of diffusion model. We apply QK-Norm \cite{henry2020query} to stabilize training and adopt the time-shift training strategy proposed in RAE \cite{zheng2025diffusion}.
More implementation details can be found in the Appendix \ref{sec:implementaion_details}.

\paragraph{Evaluation}
For reconstruction, we evaluate on the ImageNet-1k \cite{russakovsky2015imagenet} validation set and the COCO2017 \cite{lin2014microsoft} validation set at $256\times256$ resolution, reporting PSNR, SSIM, and LPIPS \cite{zhang2018unreasonable}. For generation, we report the Frechet Inception Distance(gFID) \cite{heusel2017gans}, Inception Score (IS) \cite{salimans2016improved}, as well as Precision and Recall. These metrics are computed using 50K generated images, except for ablation studies, where 10K samples are used. Additionally, we report the Frechet Distance computed on DINOv2 \cite{oquab2023dinov2} features (FDD) to complement FID, as FDD is a more reliable metric \cite{stein2023exposing, skorokhodov2025improving}.

\subsection{Ablations}

\paragraph{Effect of LV-RAE compared with fine-tuning the VFM.} \label{exp:ab_ft}
To assess the effectiveness of the proposed LV-RAE, we conduct experiments using different VFMs, including DINOv2-B~\cite{oquab2023dinov2}, DINOv3-B~\cite{simeoni2025dinov3}, and SigLIPv2-B~\cite{tschannen2025siglip}. We primarily compare reconstruction quality and semantic alignment with VFM features. Specifically, reconstruction performance is evaluated using PSNR and SSIM on the COCO2017 validation set. To measure semantic alignment, we compute CKNNA on the first 5,000 images from the ImageNet validation set. The quantitative results are summarized in Tab.~\ref{tab:ab_lvrae}. In our experiments, F.T. VFM denotes directly fine-tuning the VFM as the encoder, while Init D. indicates whether decoder initialization is applied, i.e., the decoder is first trained with a frozen VFM encoder and the VFM is then unfrozen for joint fine-tuning after convergence. 

Compared with directly fine-tuning the VFM, LV-RAE consistently achieves superior performance in both reconstruction fidelity and semantic alignment. We further observe that different VFMs exhibit distinct behaviors. In particular, compared with DINOv2 and SigLIPv2, DINOv3 provides semantic features that retain richer low-level information. As a result, both direct VFM fine-tuning and LV-RAE benefit from stronger reconstruction performance and improved semantic alignment when using DINOv3 as the backbone. We therefore adopt it as the default VFM in our experiments.

To further analyze the source of LV-RAE’s superiority, we illustrate the evolution of training losses in Fig.~\ref{fig:train_loss}. Compared with directly fine-tuning the VFM, LV-RAE exhibits a smaller and more stable alignment loss throughout training. We also conduct an ablation study on the decoder parameter initialization, and observe that a well-designed initialization slightly improves reconstruction fidelity, while having a negligible impact on semantic alignment and training stability. We attribute the performance gains of LV-RAE to its fixed semantic space, which enables stable optimization. In contrast, fine-tuning the VFM continuously alters the semantic space, leading to unstable alignment optimization and degraded reconstruction.
\vspace{-2mm}
\paragraph{Effect of noise augmentation for reconstruction}
We investigate the reconstruction ability of different decoders under varying levels of latent corruption, with quantitative results reported in Tab.~\ref{tab:noise4rec}. When evaluated on clean latent $z$, the decoder without fine-tuning achieves the best reconstruction performance. However, its performance degrades sharply even under mild noise perturbations ($z + 0.1\epsilon$), indicating limited robustness to latent corruption. In contrast, the fine-tuned decoder exhibits a substantial improvement in robustness, maintaining significantly better reconstruction quality as the noise level increases. We further compare fine-tuning with a fixed noise magnitude ($\sigma=0.1$) and fine-tuning with randomly sampled noise amplitudes ($\sigma\sim \mathcal{U}(0, 0.2)$). The results show that fine-tuning with a fixed noise level severely harms reconstruction performance on clean latent. We attribute this degradation to an implicit high-frequency truncation effect, where the decoder no longer attends to fine-grained latent information. In comparison, fine-tuning with random noise amplitudes enhances robustness while largely preserving reconstruction quality on clean latent inputs. We therefore adopt LV-RAE equipped with a fine-tuned decoder using $\sigma\sim\mathcal{U}(0,0.2)$ as the default setting.

\begin{table}[!t]
    \centering
    \caption{\textbf{Quantitative comparison of different decoders under increasing levels of latent corruption.} The original decoder achieves the best reconstruction performance on clean latents but degrades sharply when noise is introduced. Fine-tuning with a fixed noise magnitude improves robustness at the cost of reconstruction quality on clean latents, while fine-tuning with randomly sampled noise amplitudes preserves high reconstruction quality on clean latents and significantly enhances robustness to noise.}
    \resizebox{\linewidth}{!}{
    \begin{tabular}{lcccc}
    \toprule
    Decoder & Latent & PSNR$\uparrow$ & SSIM$\uparrow$ & LPIPS$\downarrow$ \\
    \midrule
    \multirow{3}{*}{original} & $z$ & 32.32 & 0.941 & 0.015\\
     & $z+0.1\epsilon$ & 17.72 & 0.437 & 0.466 \\
     & $z+0.2\epsilon$ & 13.68 & 0.228 & 0.621 \\
    \midrule
    \multirow{3}{*}{F.T. w/ $\sigma=0.1$ } & $z$ & 25.45 &  0.810 & 0.092 \\
     & $z+0.1\epsilon$ & 23.56 & 0.735 & 0.093 \\
     & $z+0.2\epsilon$ & 19.98 & 0.581 & 0.182 \\
    \midrule
    \multirow{3}{*}{F.T. w/ $\sigma\sim\mathcal{U}(0, 0.2)$ } & $z$ & 30.91  & 0.930 & 0.018\\
     & $z+0.1\epsilon$ & 23.78 & 0.735 & 0.100 \\
     & $z+0.2\epsilon$ & 21.61 & 0.637 & 0.152 \\
    \bottomrule
    \end{tabular}
    }
    
    \label{tab:noise4rec}
\end{table}

\begin{table}[t]
    \centering
    \caption{\textbf{Additional quantitative comparison of generation performance.} LV-RAE achieves a gFDD of 58.2 without guidance, substantially outperforming other methods. $\dagger$ Results reproduced using the official open-source weights.}
    \resizebox{\linewidth}{!}{
    \begin{tabular}{lcccc}
    \toprule
       Method  &  Epochs & Steps & gFID$\downarrow$  & gFDD$\downarrow$  \\
    \midrule
       SVG$\dagger$ \cite{shi2025latent}  & 1400 & 250 & 3.47  & 130.4 \\
       VA-VAE$\dagger$ \cite{yao2025reconstruction} & 800 & 250 &\textbf{2.23} & 74.3 \\
       \rowcolor{gray!15}
       LV-RAE (\textcolor{blue!70}{ours}) & 800 & 250 & 2.42 &\textbf{58.2} \\ 
    \bottomrule
    \end{tabular}
    }
    \label{tab:g_FDD}
    \vspace{-4mm}
\end{table}

\begin{figure}[t]
    \centering
    \includegraphics[width=\linewidth]{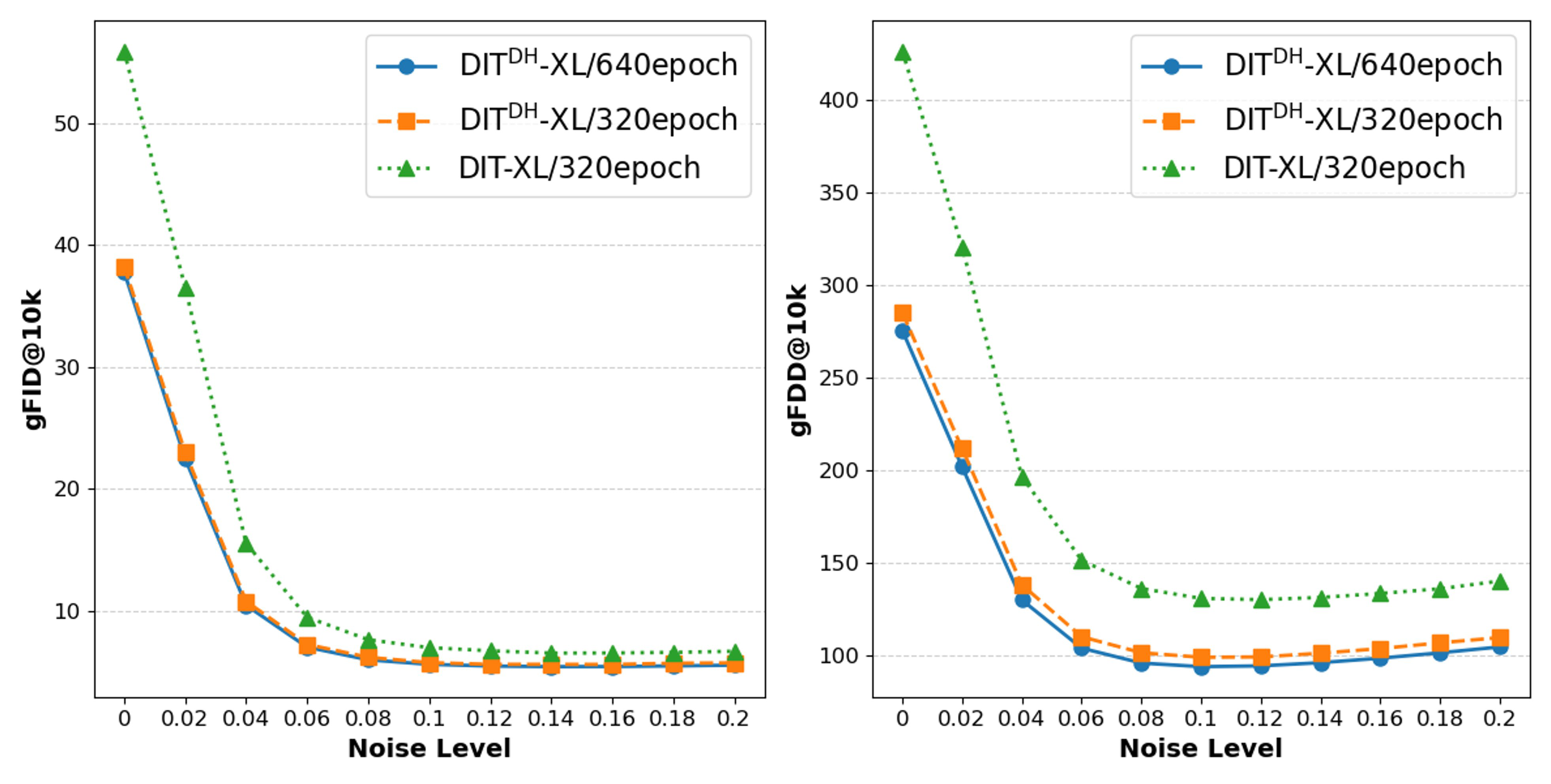}
    \caption{\textbf{Generation quality trends under different levels of noise injection at inference time.} As the noise level increases, both FID and FDD decrease substantially at first and then exhibit a slight increase.
    }
    \label{fig:noise_effect}

\end{figure}

\paragraph{Effect of noise augmentation for generation}\label{sec:ab_noise_for_generation}
We add noise to the final sampled latent representations to suppress visual artifacts caused by inaccurate generation of high-frequency components. Fig.~\ref{fig:noise_effect} visualizes the effect of different noise levels on the final generation quality. As the noise magnitude increases, the generation quality improves consistently: both FID and FDD decrease with increasing noise strength and begin to converge at around $\bar{\sigma} \approx 0.08$.
We further compare models trained with different numbers of epochs (320 vs.\ 640) and observe that longer training has a limited impact on the overall trend, mainly resulting in a vertical shift of the curves. In contrast, the choice of model architecture has a much more significant effect. Specifically, when the noise level is low, larger-capacity models (e.g., $\mathrm{DiT}^{\mathrm{DH}}$-XL) achieve substantially better performance than their smaller counterparts (e.g., DiT-XL).

\subsection{Reconstruction Performance Comparison}
The quantitative reconstruction results are summarized in Tab.~\ref{tab:reconstruction_result}. Compared with other tokenizers, the latents produced by LV-RAE preserve richer fine-grained details, leading to state-of-the-art reconstruction performance. After applying the proposed noise augmentation training strategy to fine-tune the decoder, reconstruction performance decreases but remains better than that of other tokenizers. Qualitative results are shown in Fig.~\ref{fig:rec_result_vis}. Previous representation autoencoders (e.g., SVG) suffer from severe structural distortions in the reconstructed images, which significantly degrade visual quality. In contrast, LV-RAE is able to handle these challenging scenarios effectively, producing structurally coherent and visually faithful reconstructions.

\subsection{Generation Performance Comparison}
As shown in Tab.~\ref{tab:generation_result}, our method achieves strong class-conditional generation performance on ImageNet 256$\times$256, outperforming most latent diffusion baselines. With 400 training epochs, DiT-XL equipped with LV-RAE attains a gFID of 3.77 and an IS of 185.37 without classifier guidance. When scaling the model to $\text{DiT}^{\text{DH}}$-XL, our method achieves a gFID of 2.42 and an IS of 223.84 after 800 training epochs, which is competitive with or superior to existing latent diffusion methods of comparable scale. We further compare performance under the FDD metric, with results reported in Tab.~\ref{tab:g_FDD}. Our method achieves an FDD score of 58.2 after 800 training epochs, substantially outperforming prior approaches such as VA-VAE and SVG.

\section{Conclusion}
In this paper, we revisit the reconstruction bottleneck that arises when using VFMs as autoencoders. To address this limitation, we propose an improved representation autoencoder, LV-RAE, which augments semantic features with missing low-level information. This design enables high-fidelity reconstruction while remaining highly aligned with the semantic distribution.
Moreover, we find that in high-dimensional spaces, the decoder is highly sensitive to off-manifold directions, which significantly hinders generation performance. To mitigate this issue, we enhance the decoder’s robustness by fine-tuning it with noise, and during inference, we inject noise to smooth the generated latent, leading to improved generation quality.
Overall, our approach effectively alleviates the reconstruction bottleneck and improves the stability and quality of generation.

\section*{Impact Statement}
This paper improves the autoencoders used in latent diffusion models, focusing on the simultaneous enhancement of reconstruction quality and semantic distribution. These advancements support diverse applications in creative content generation and generative AI research. Our method does not expand the functional domain of existing models; therefore, it inherits the standard ethical considerations and societal challenges associated with large-scale generative systems.

\bibliography{main}
\bibliographystyle{icml2026}

\newpage
\appendix
\onecolumn
\clearpage
\setcounter{page}{1}
\setcounter{section}{0}
\renewcommand\thesection{\Alph{section}}

\section{Background}
\paragraph{Flow Matching}
Flow matching \cite{liu2022flow,lipman2022flow,esser2024scaling} learns a deterministic 
probability flow that continuously transports a known prior distribution 
$p_1=\mathcal{N}(0,\mathrm{I})$ to the target data distribution $p_0$. To estimate a 
time-dependent velocity field $v_t$ that induces a probability path between $p_0$ and $p_1$, 
we can define a forward interpolating process between $x_0 \sim p_0$ and 
$\epsilon \sim p_1$:
\begin{equation}
    x_t = \alpha_{t} x_0 + \beta_t \epsilon, \quad 0 \le t \le 1,
\end{equation}
where the coefficients satisfy the boundary conditions 
$\alpha_0=\beta_1=1$ and $\alpha_1=\beta_0=0$.

The distribution of this stochastic interpolant coincides with that of the 
solution to a deterministic ODE whose drift is the corresponding conditional expectation. 
Let
$
v_t(X) = \mathbb{E}\!\left[\frac{d x_t}{d t}\,\middle|\,x_t=X\right],
$
then the probability flow ODE
\begin{equation}
    dX_t = v_t(X_t)dt, \quad X_1 \sim p_1
\end{equation}
satisfy 
$\mathrm{Law}(X_t)=\mathrm{Law}(x_t)$ for all $t\in[0,1]$.

To learn the velocity field $v_t$, we can use empirical risk minimization to 
approximate this conditional expectation in practice.  This is achieved by optimizing the square loss function:
\begin{equation}
    \mathcal{L}(\theta)
    =
    \mathbb{E}_{t,x_0,\epsilon}
    \left[
    \|v_{\theta}(x_t,t)-\dot{x}_t\|^2_{2}
    \right],
\end{equation}
where $\dot{x}_t=\frac{d x_t}{d t}$ is analytically available.

\section{Implementation Details}
\label{sec:implementaion_details}

\subsection{Diffusion Model}
\paragraph{DiT and $\text{DiT}^{\text{DH}}$ model details.}
Our DiT and $\text{DiT}^{\text{DH}}$ models are built upon the RAE~\cite{zheng2025diffusion} configuration. We adopt SwiGLU~\cite{shazeer2020glu} activations and RMSNorm~\cite{zhang2019root}, together with a Gaussian Fourier embedding~\cite{song2020score} layer for timestep encoding. For positional encoding, we apply Absolute Positional Embeddings to the input tokens in addition to RoPE~\cite{su2024roformer}. To improve training stability, we incorporate QK-Norm~\cite{henry2020query}; while QK-Norm slightly affects final generation performance, it provides more stable and reliable optimization during training~\cite{yao2025reconstruction, chen2025aligning}.

We follow prior work~\cite{Peebles2022DiT, zheng2025diffusion} and use the same model sizes for DiT-XL and $\text{DiT}^{\text{DH}}$-XL. We use $\text{DiT}^{\text{DH}}$-S as the guiding model for AutoGuidance\cite{karras2024guiding}:
\begin{itemize}
\item DiT-XL: hidden dimensionality of 1152, 28 transformer blocks, and 16 attention heads.
\item DiT-S: hidden dimensionality of 384, 12 transformer blocks, and 6 attention heads.
\item $\text{DiT}^{\text{DH}}$-XL: DiT-XL augmented with a DDT head (consisting of a hidden dimensionality of 2048, 2 transformer blocks, and 16 attention heads.)
\item $\text{DiT}^{\text{DH}}$-S: DiT-S augmented with a DDT head (consisting of a hidden dimensionality of 2048, 2 transformer blocks, and 16 attention heads.)
\end{itemize}

\paragraph{DiT and $\text{DiT}^{\text{DH}}$ training details.}
All DiT and $\text{DiT}^{\text{DH}}$ models are trained on NVIDIA A100 40GB GPUs using mixed-precision training with bfloat16. We use the AdamW\cite{loshchilov2017decoupled} optimizer with a learning rate of $2.0 \times 10^{-4}$ for the first 40 epochs, which is then decayed to $2.0 \times 10^{-5}$ until epoch 800. The total batch size is set to 1024. Following~\cite{zheng2025diffusion}, we apply EMA of model weights with a decay rate of 0.9995.

\paragraph{Time shifting.}
We adopt the time-shifting schedule from RAE~\cite{zheng2025diffusion} for both training and sampling:
\begin{equation}
    t' = \frac{\alpha \cdot t}{1 + (\alpha - 1)\cdot t}, \quad 
    \alpha = \sqrt{\frac{c \cdot h \cdot w}{4096}} .
\end{equation}
Here, $t$ denotes the diffusion timestep, ranging from 1 (full noise) to 0 (clean image), and $\alpha$ is the shifting coefficient. For a $256\times256$ input image, our autoencoder encodes it into a latent representation with $h=16$, $w=16$, and $c=768$, resulting in $\alpha \approx 6.93$.

\paragraph{Sampling.}
We use standard ODE-based sampling with the Euler sampler, employing 250 sampling steps by default. Following~\cite{zheng2025diffusion}, we adopt AutoGuidance as our primary guidance method. We train the smallest variant, $\text{DiT}^{\text{DH}}$-S, for 16 epochs as the guiding model and use a guidance scale of 1.4.

\subsection{Autoencoders}
\paragraph{LV-RAE model details.}
LV-RAE is implemented as a Transformer-based autoencoder equipped with RoPE~\cite{su2024roformer} and T5-MLP\cite{raffel2020exploring}. The model configuration is as follows:
\begin{itemize}
    \item Encoder: hidden dimensionality of 768, 6 transformer blocks, and 12 attention heads.
    \item Decoder: hidden dimensionality of 768, 12 transformer blocks, and 12 attention heads.
\end{itemize}

\paragraph{LV-RAE training details.}
During the local-variations augmented stage (training stage I), each autoencoder is trained using the AdamW optimizer~\cite{loshchilov2017decoupled} with $\beta_1=0.9$, $\beta_2=0.999$, and a weight decay of 0.01. We use a constant learning rate of $1.0 \times 10^{-4}$ and a total batch size of 512. Each autoencoder is trained for 80k steps.

In the noise augmentation stage (training stage II), the encoder is frozen and the decoder is fine-tuned using the same optimization settings as in the previous stage. 
We first train the decoder using only the reconstruction loss for 10k steps, and then introduce the GAN loss for an additional 90k training steps.
Following RAE~\cite{zheng2025diffusion}, we use DINO-S/8 as the discriminator. We additionally apply differentiable augmentations~\cite{zhao2020differentiable}, followed by a random crop to $224\times224$ before feeding samples into the discriminator.

\section{Evaluation Details}
We evaluate our method from both reconstruction quality and generative performance perspectives, using standard metrics widely adopted in prior work.

\subsection{Standard Feature Alignment Metrics}
We use \textbf{CKNNA}(Centered Kernel Nearest-Neighbor Alignment)\cite{huh2024platonic} to evaluate the alignment between the latent representations learned by LV-RAE and the semantic features extracted from VFM.
CKNNA is a relaxed version of Centered Kernel Alignment\cite{kornblith2019similarity} that measures representation alignment by comparing local neighborhood structures. We follow the evaluation protocol of the original work\footnote{https://github.com/minyoungg/platonic-rep} for its computation.

\subsection{Standard Reconstruction Metrics}
To assess reconstruction fidelity, we report \textbf{PSNR}, \textbf{SSIM}, and \textbf{LPIPS}.
\begin{itemize}
    \item \textbf{PSNR} measures pixel-wise reconstruction accuracy and is sensitive to low-level differences. 
    \item \textbf{SSIM} evaluates structural similarity between reconstructed images and ground truth, emphasizing luminance, contrast, and structural consistency.
    \item \textbf{LPIPS} measures perceptual similarity using deep features from pretrained networks and correlates well with human perception. 
\end{itemize}

\subsection{Standard Generative Metrics}
To evaluate generative performance, we report \textbf{FID}, \textbf{IS}, \textbf{Precision} and \textbf{Recall}. We strictly follow the setup and use the same reference batches of ADM\footnote{https://github.com/openai/guided-diffusion}\cite{dhariwal2021diffusion} for evaluation.

\begin{itemize}
    \item \textbf{FID} measures the distance between the distributions of generated and real images in the feature space of the Inception-v3 network~\cite{szegedy2016rethinking}.
    \item \textbf{IS} evaluates both sample quality and diversity by measuring the confidence and entropy of class predictions. 
    \item \textbf{Precision} and \textbf{Recall} explicitly disentangle fidelity and diversity: precision measures the quality of generated samples, while recall reflects coverage of the real data distribution.
\end{itemize}

\subsection{Frechet Distance computed on top of DINOv2 features (FDD)}

In addition to standard FID, we report FDD, a feature distribution distance computed using DINOv2 features. Specifically, we replace Inception features with class token extracted from a pretrained DINOv2 model\footnote{https://huggingface.co/facebook/dinov2-large} and compute the Fréchet distance in this semantic feature space.

Compared to Inception-based FID, DINO-based FID correlates better with human perceptual judgments\cite{stein2023exposing}, and thus provides a more appropriate measure for evaluating image distributions.

\section{Additional Qualitative Results}

\subsection{PCA Visualizations}

\begin{figure}[h]
    \centering
    \includegraphics[width=\textwidth]{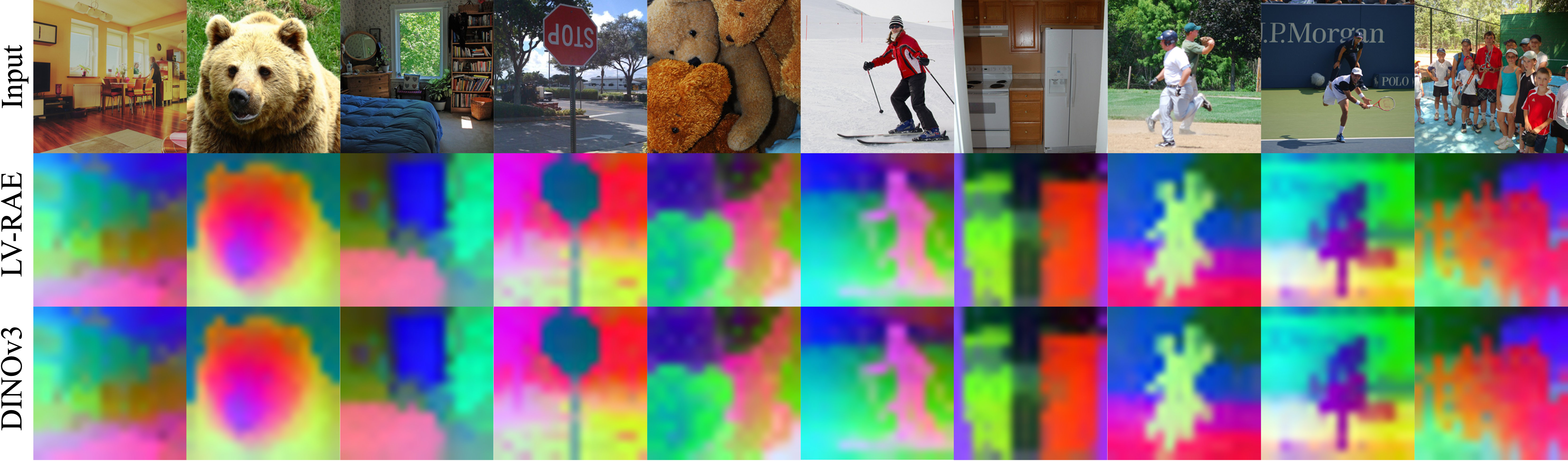}
    \caption{
    \textbf{PCA visualizations of LV-RAE latent and DINOv3 semantic features.} The two visualizations exhibit strong consistency, indicating a high degree of semantic alignment between LV-RAE and DINOv3.
    }
    \label{fig:pca_results}
\end{figure}

Fig.~\ref{fig:pca_results} shows the PCA \cite{hotelling1933analysis} visualizations of the features extracted by LV-RAE and DINOv3.

\subsection{Effect of the LV-RAE Encoder}

\begin{figure}
    \centering
    \includegraphics[width=\linewidth]{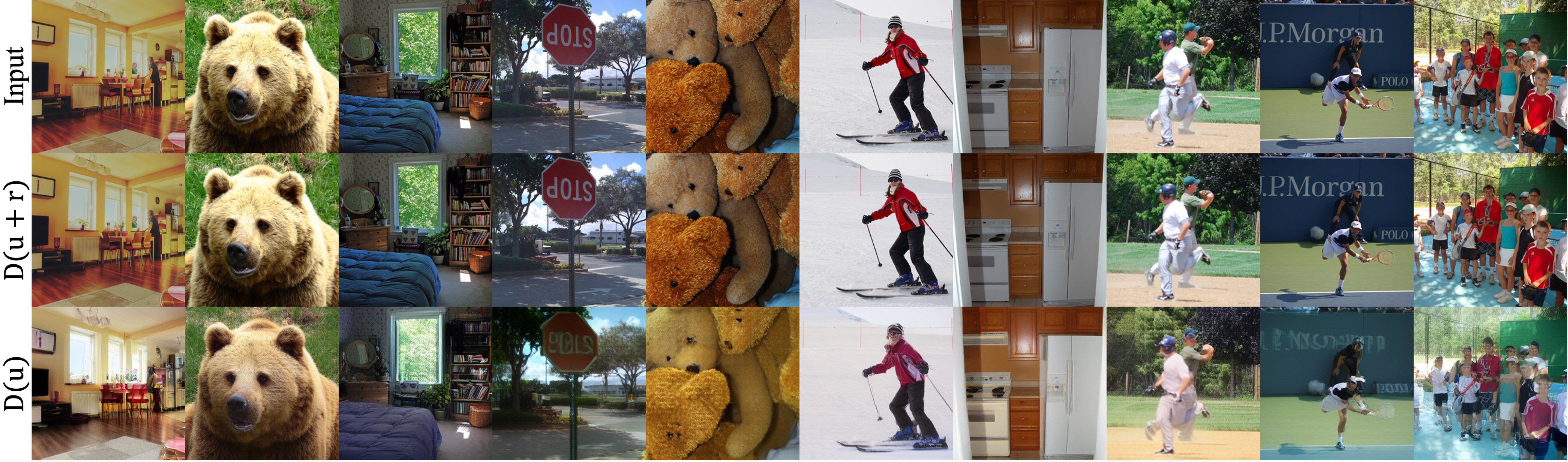}
    \caption{\textbf{Visualization of LV-RAE reconstructions under different latent inputs.} Excluding the encoder output $\textbf{r}$ leads to color distortions and missing texture details.}
    \label{fig:rec_dino}
\end{figure}

\begin{figure}
    \centering
    \includegraphics[width=\linewidth]{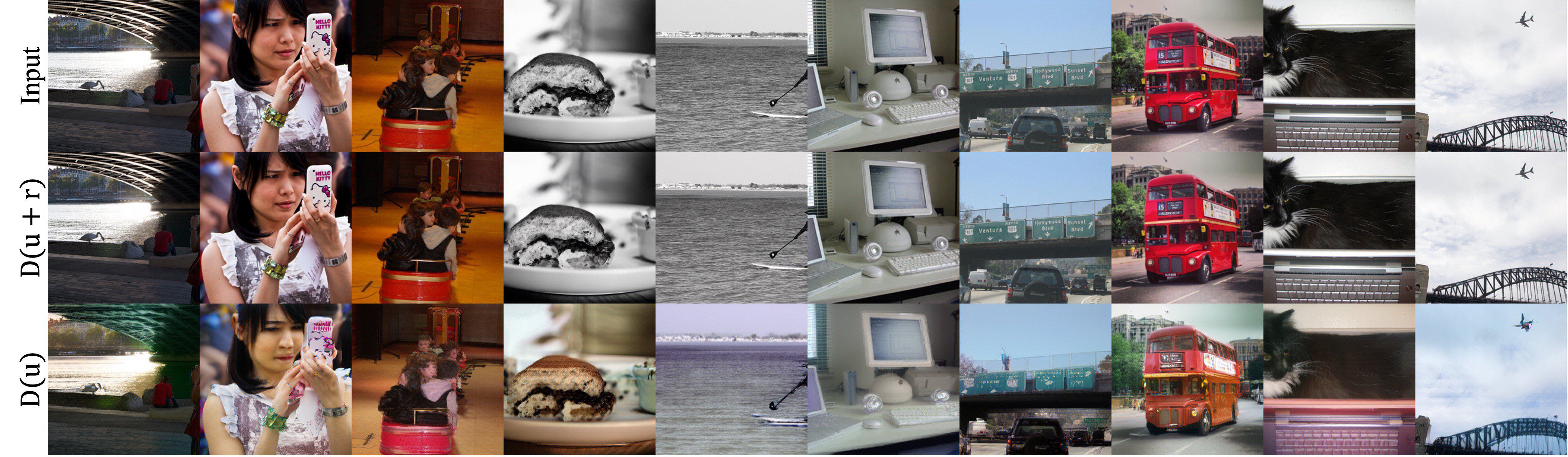}
    \caption{\textbf{Visualization of LV-RAE reconstructions under different latent inputs.} Excluding the encoder output $\textbf{r}$ leads to color distortions and missing texture details.}
    \label{fig:rec_dino_2}
\end{figure}

\begin{figure}
    \centering
    \includegraphics[width=\linewidth]{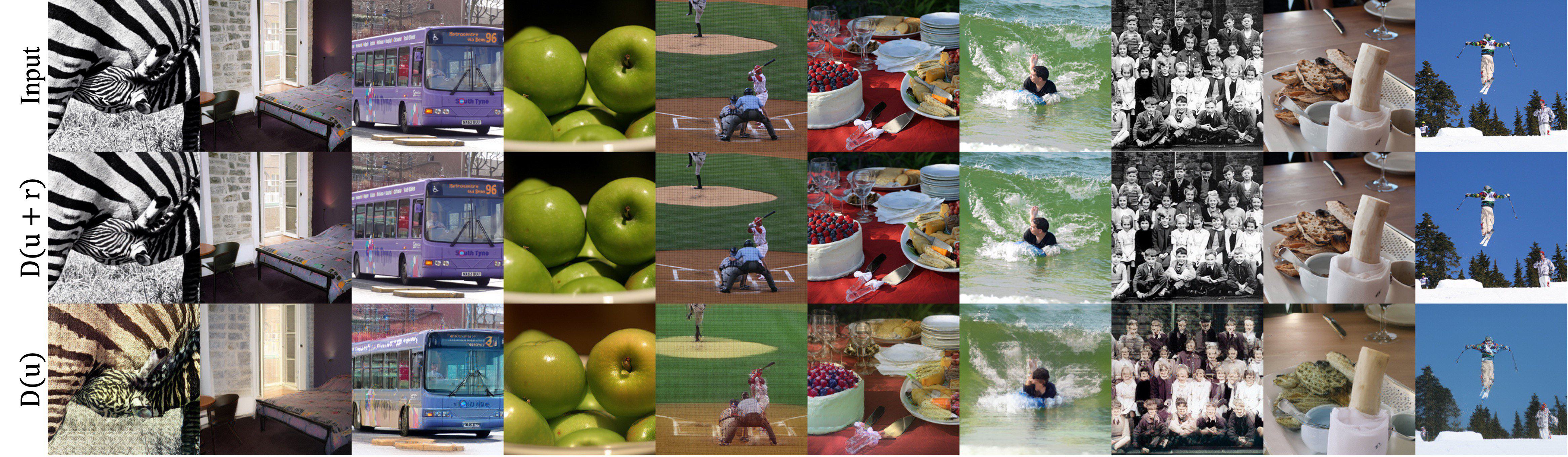}
    \caption{\textbf{Visualization of LV-RAE reconstructions under different latent inputs.} Excluding the encoder output $\textbf{r}$ leads to color distortions and missing texture details.}
    \label{fig:rec_dino_3}
\end{figure}

To verify that the LV-RAE encoder output $\textbf{r}$ captures the low-level information missing from the semantic features $\textbf{u}$, we visualize reconstructions produced by the LV-RAE decoder under different latent inputs. As shown in ~\cref{fig:rec_dino,fig:rec_dino_2,fig:rec_dino_3}, when the encoder output $\textbf{r}$ is not included, the reconstructed images exhibit noticeable color shifts and lack correct fine-grained textures. This observation indicates that the LV-RAE encoder effectively learns complementary low-level details that are critical for accurate image reconstruction.

\subsection{Effect of Noise Level on Generation Results}
\begin{figure}
    \centering
    \includegraphics[width=\linewidth]{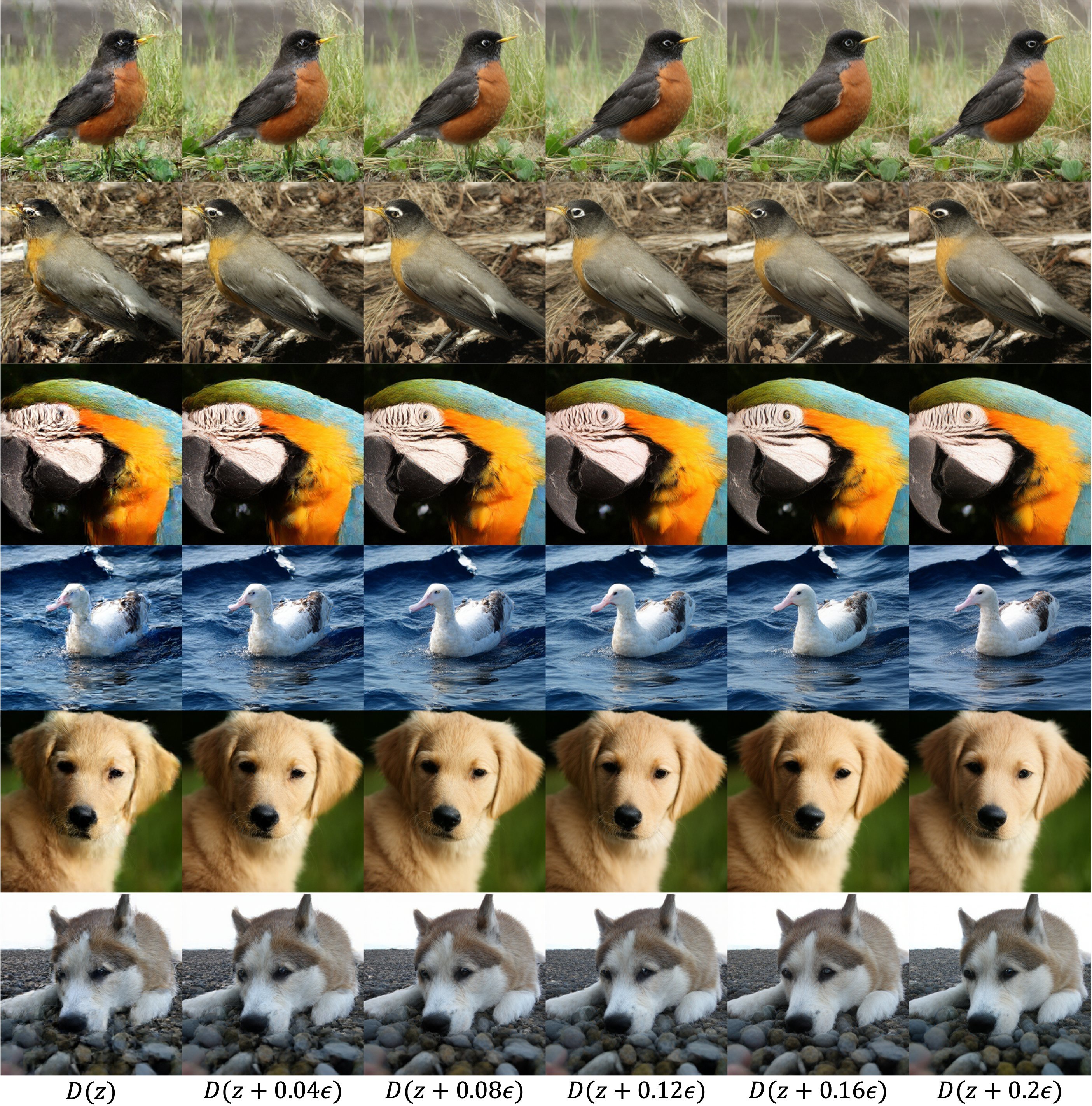}
    \caption{\textbf{Visualization of the effect of latent noise injection.} }
    \label{fig:noise_effect_1}
\end{figure}

\begin{figure}
    \centering
    \includegraphics[width=\linewidth]{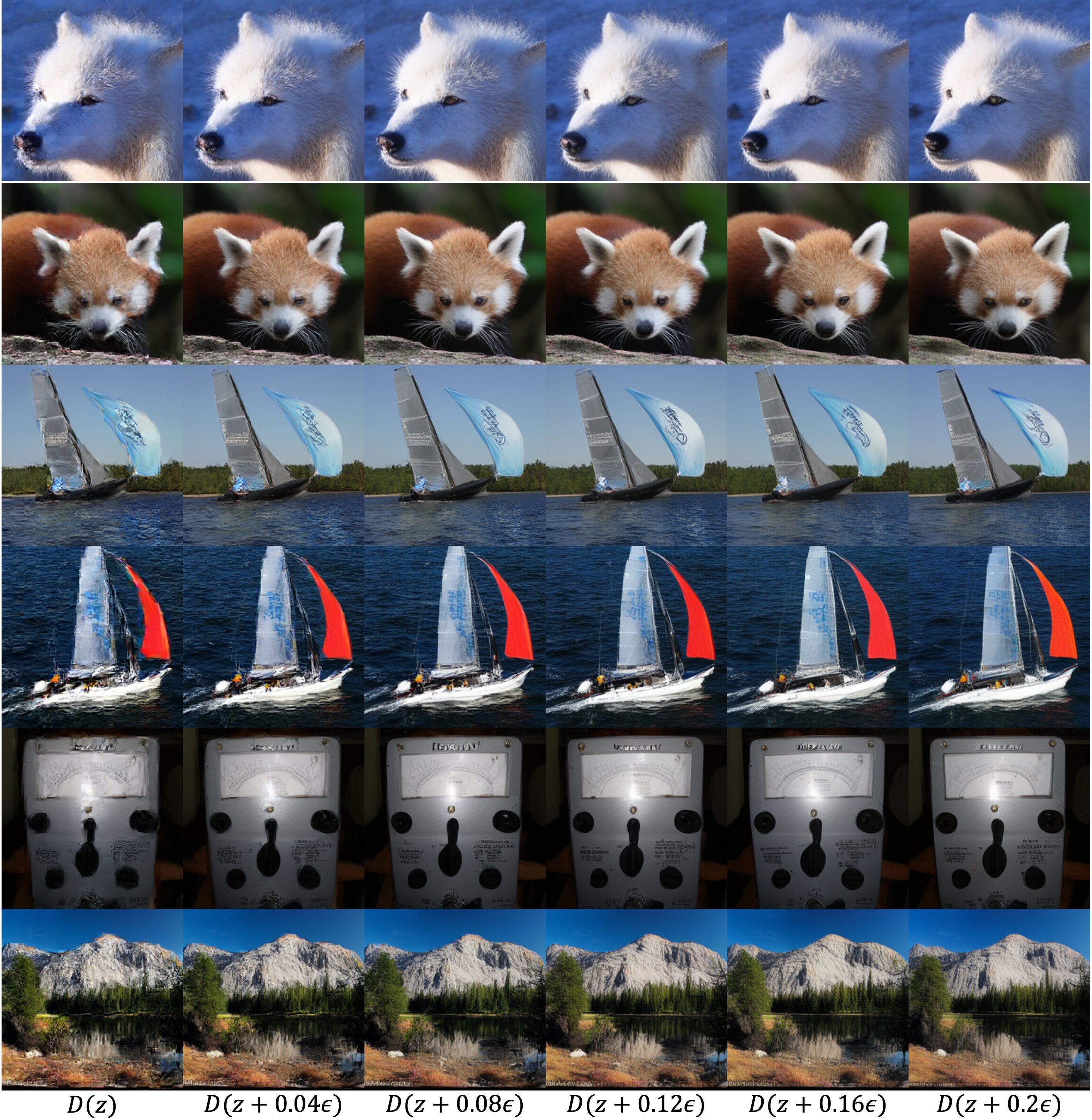}
    \caption{\textbf{Visualization of the effect of latent noise injection.}}
    \label{fig:noise_effect_2}
\end{figure}

To improve generation quality, we inject noise into the generated latent and provide visualizations illustrating the effect of different noise strengths, as shown in Figs.~\ref{fig:noise_effect_1} and~\ref{fig:noise_effect_2}.

\subsection{Uncurated Generation Visual Results}

\begin{figure}[t]
    \centering
    \includegraphics[width=0.9\linewidth]{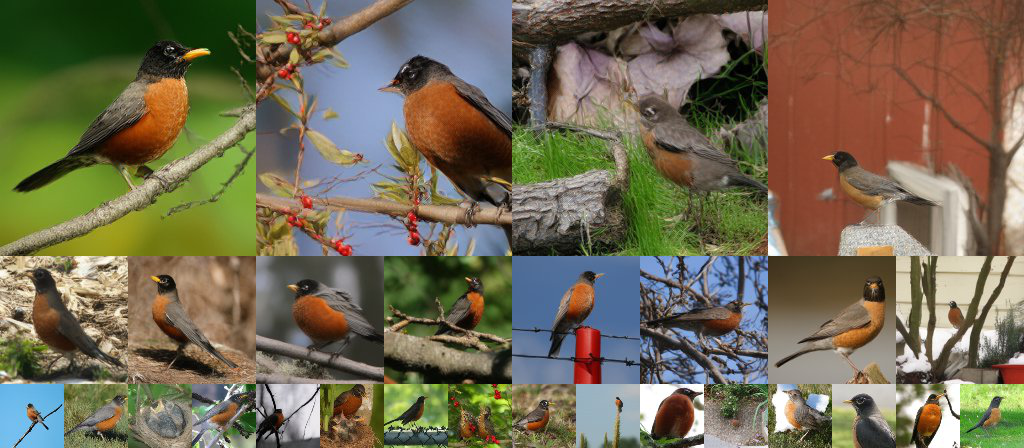}
    \caption{\textbf{Uncurated $256\times256$ $\text{DiT}^{\text{DH}}$-XL samples.} AutoGudance Scale=1.4, Class label=15}
    \label{fig:gen15}
\end{figure}

\begin{figure}[t]
    \centering
    \includegraphics[width=0.9\linewidth]{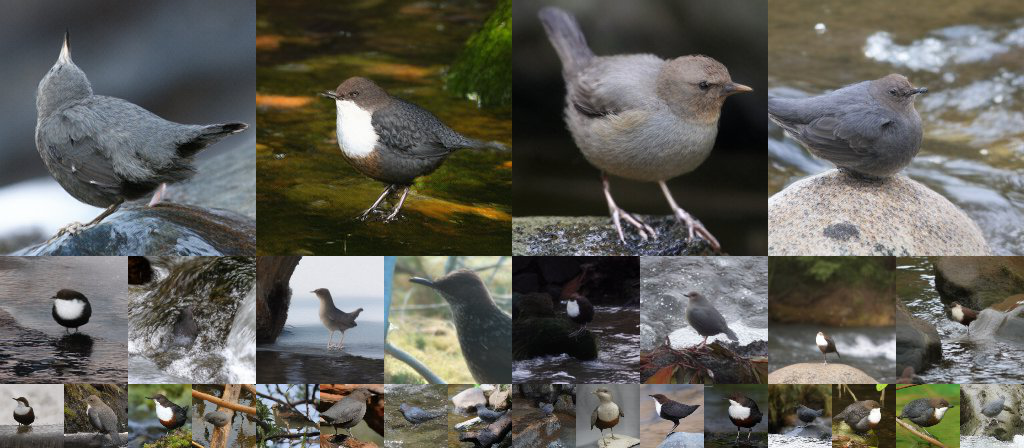}
    \caption{\textbf{Uncurated $256\times256$ $\text{DiT}^{\text{DH}}$-XL samples.} AutoGudance Scale=1.4, Class label=20}
    \label{fig:gen20}
\end{figure}

\begin{figure}[t]
    \centering
    \includegraphics[width=0.9\linewidth]{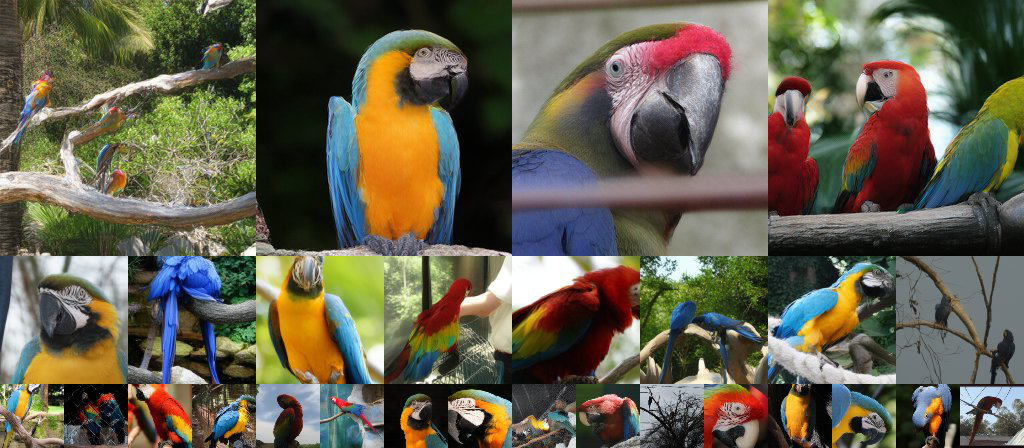}
    \caption{\textbf{Uncurated $256\times256$ $\text{DiT}^{\text{DH}}$-XL samples.} AutoGudance Scale=1.4, Class label=88}
    \label{fig:gen88}
\end{figure}

\begin{figure}[h]
    \centering
    \includegraphics[width=0.9\linewidth]{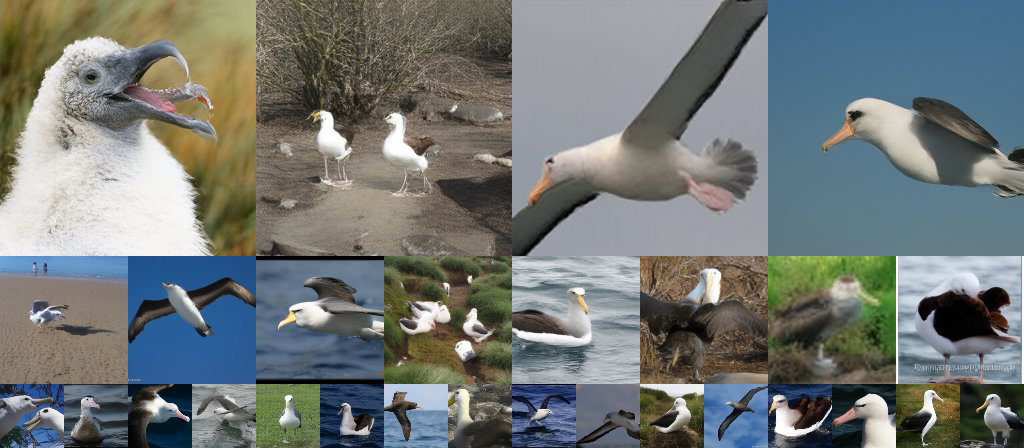}
    \caption{\textbf{Uncurated $256\times256$ $\text{DiT}^{\text{DH}}$-XL samples.} AutoGudance Scale=1.4, Class label=146}
    \label{fig:gen146}
\end{figure}

\begin{figure}[h]
    \centering
    \includegraphics[width=0.9\linewidth]{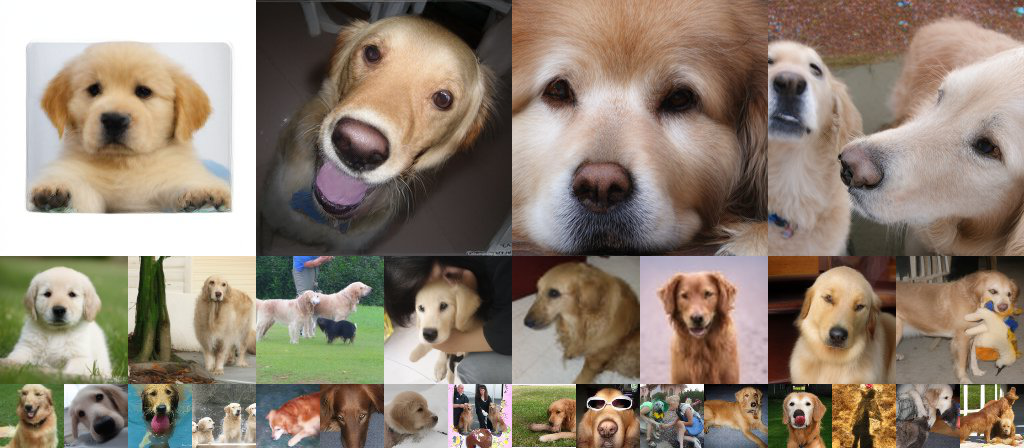}
    \caption{\textbf{Uncurated $256\times256$ $\text{DiT}^{\text{DH}}$-XL samples.} AutoGudance Scale=1.4, Class label=207}
    \label{fig:gen207}
\end{figure}

\begin{figure}[h]
    \centering
    \includegraphics[width=0.9\linewidth]{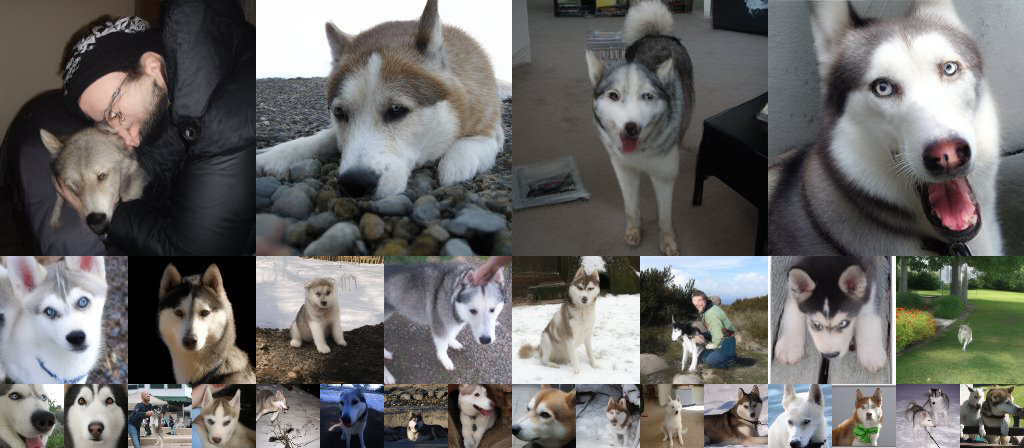}
    \caption{\textbf{Uncurated $256\times256$ $\text{DiT}^{\text{DH}}$-XL samples.} AutoGudance Scale=1.4, Class label=250}
    \label{fig:gen250}
\end{figure}

\begin{figure}[h]
    \centering
    \includegraphics[width=0.9\linewidth]{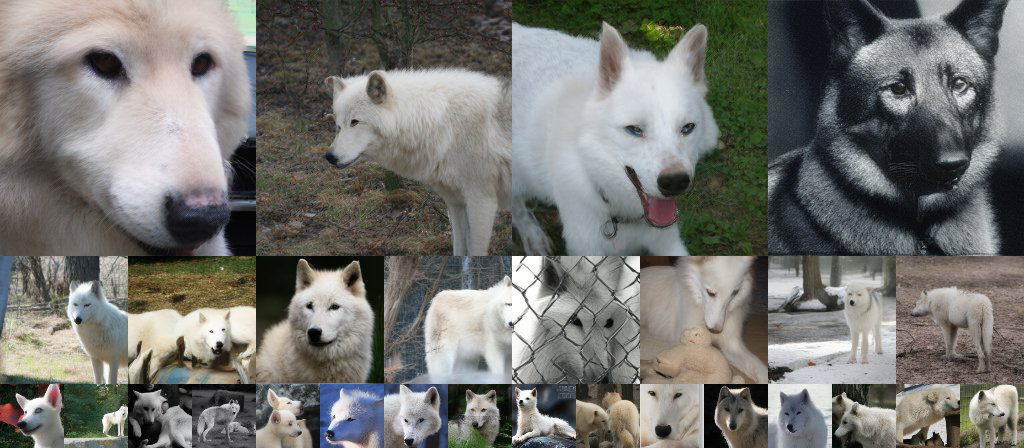}
    \caption{\textbf{Uncurated $256\times256$ $\text{DiT}^{\text{DH}}$-XL samples.} AutoGudance Scale=1.4, Class label=270}
    \label{fig:gen270}
\end{figure}

\begin{figure}[h]
    \centering
    \includegraphics[width=0.9\linewidth]{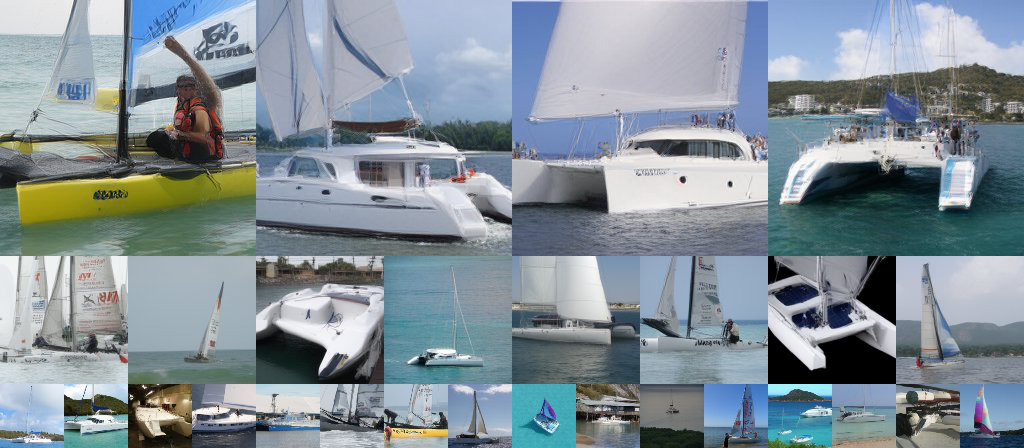}
    \caption{\textbf{Uncurated $256\times256$ $\text{DiT}^{\text{DH}}$-XL samples.} AutoGudance Scale=1.4, Class label=484}
    \label{fig:gen484}
\end{figure}

\begin{figure}[h]
    \centering
    \includegraphics[width=0.9\linewidth]{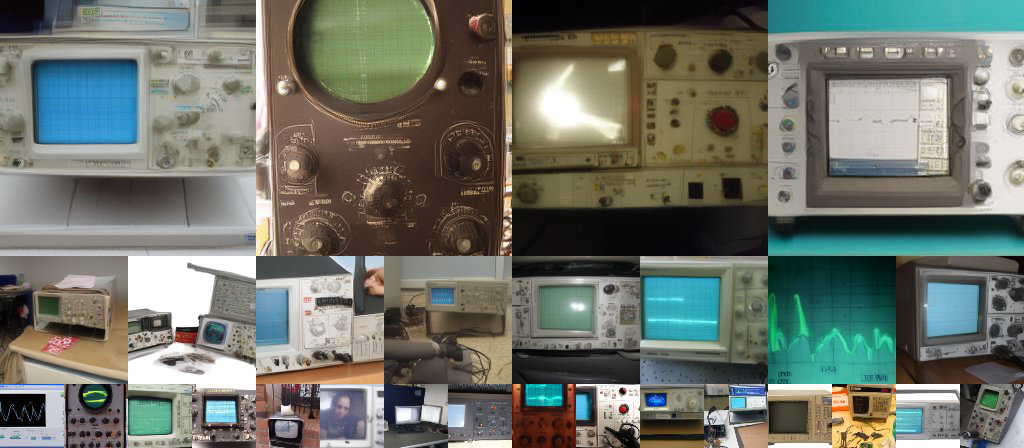}
    \caption{\textbf{Uncurated $256\times256$ $\text{DiT}^{\text{DH}}$-XL samples.} AutoGudance Scale=1.4, Class label=688}
    \label{fig:gen688}
\end{figure}

\begin{figure}[h]
    \centering
    \includegraphics[width=0.9\linewidth]{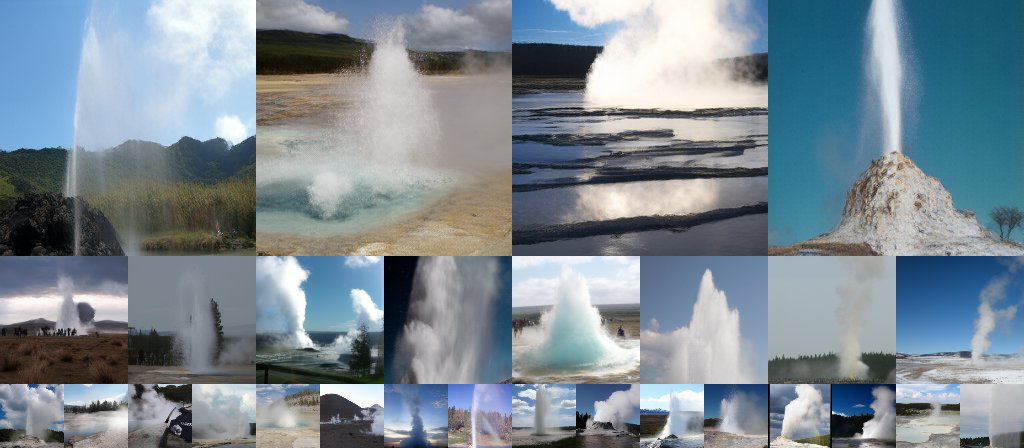}
    \caption{\textbf{Uncurated $256\times256$ $\text{DiT}^{\text{DH}}$-XL samples.} AutoGudance Scale=1.4, Class label=974}
    \label{fig:gen974}
\end{figure}

\begin{figure}[h]
    \centering
    \includegraphics[width=0.9\linewidth]{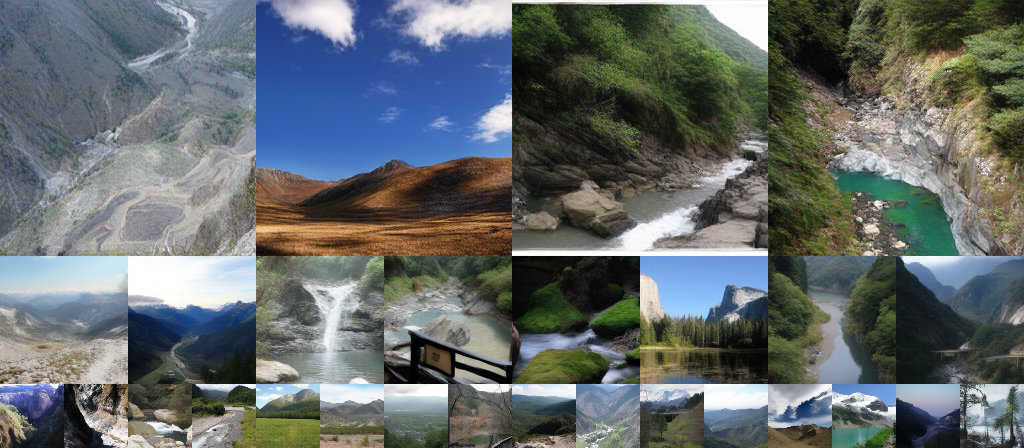}
    \caption{\textbf{Uncurated $256\times256$ $\text{DiT}^{\text{DH}}$-XL samples.} AutoGudance Scale=1.4, Class label=979}
    \label{fig:gen979}
\end{figure}

\begin{figure}[h]
    \centering
    \includegraphics[width=0.9\linewidth]{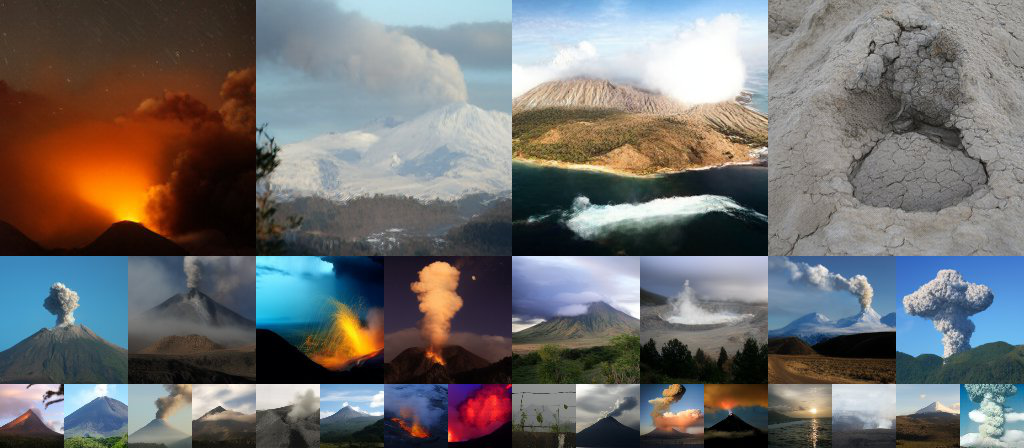}
    \caption{\textbf{Uncurated $256\times256$ $\text{DiT}^{\text{DH}}$-XL samples.} AutoGudance Scale=1.4, Class label=980}
    \label{fig:gen980}
\end{figure}

We provide uncurated generation results for specific classes in \cref{fig:gen15,fig:gen20,fig:gen146,fig:gen207,fig:gen250,fig:gen270,fig:gen484,fig:gen688,fig:gen88,fig:gen974,fig:gen979,fig:gen980}.

\end{document}